\documentclass{article} 
\usepackage{iclr2025_conference,times}

\usepackage{amsmath,amsfonts,bm}

\def\eqref#1{equation~\ref{#1}}

\def\1{\bm{1}}

\DeclareMathAlphabet{\mathsfit}{\encodingdefault}{\sfdefault}{m}{sl}
\SetMathAlphabet{\mathsfit}{bold}{\encodingdefault}{\sfdefault}{bx}{n}

\usepackage[colorlinks,
		linkcolor = blue,		
		anchorcolor = blue,	
		urlcolor = blue,		
		citecolor = blue,		
          ]{hyperref}
\usepackage{booktabs}       
\usepackage{graphicx}
\usepackage{amsmath}
\usepackage{cases}
\usepackage{bm}
\usepackage{amsmath}
\usepackage{bbm}
\usepackage{multirow}
\usepackage{subfigure}
\usepackage{arydshln}
\usepackage{subfigure}
\usepackage{xcolor} 
\usepackage{colortbl} 

\usepackage{algorithm}
\usepackage{listings}
\usepackage{wrapfig}
\usepackage{algpseudocode}
\usepackage{makecell}

\usepackage{listings}
\usepackage{xcolor}

\definecolor{vscodeBlue}{RGB}{97, 175, 239}
\definecolor{vscodeGreen}{RGB}{152, 195, 121}
\definecolor{vscodePurple}{RGB}{198, 120, 221}
\definecolor{vscodeOrange}{RGB}{224, 108, 117}
\definecolor{vscodeGray}{RGB}{128, 128, 128}
\newcommand{\revise}[1]{{\color{black} #1}}

\title{KaSA: Knowledge-Aware Singular-Value Adaptation of Large Language Models}

\author{Fan Wang\thanks{Equal contributors: Fan Wang, Juyong Jiang, and Chansung Park.}$~^\heartsuit$, 
Juyong Jiang\footnotemark[1]$~^\heartsuit$, 
Chansung Park\footnotemark[1]$~^\spadesuit$,
Sunghun Kim\thanks{Corresponding authors: Sunghun Kim and Jing Tang.}$^{~\heartsuit\clubsuit}$, 
Jing Tang\footnotemark[2]$^{~\heartsuit\clubsuit}$ \\ 
$^\heartsuit$The Hong Kong University of Science and Technology (Guangzhou) \\
$^\spadesuit$Electronics and Telecommunications Research Institute\\
$^\clubsuit$The Hong Kong University of Science and Technology\\
\texttt{\{csfanwang,csjuyongjiang,deep.diver.csp\}@gmail.com}\\
\texttt{\{hunkim,jingtang\}@ust.hk}
}

\iclrfinalcopy 
\begin{document}

\maketitle

\begin{abstract}
The increasing sizes of large language models (LLMs) result in significant computational overhead and memory usage when adapting these models to specific tasks or domains. Various parameter-efficient fine-tuning (PEFT) methods have been devised to mitigate these challenges by training a small set of parameters for the task-specific updates of the model weights. Among PEFT methods, LoRA stands out for its simplicity and efficiency, inspiring the development of a series of variants. However, LoRA and its successors disregard the knowledge that is noisy or irrelevant to the targeted task, detrimentally impacting model performance and leading to suboptimality. To address this limitation, we introduce Knowledge-aware Singular-value Adaptation (KaSA), a PEFT method that leverages singular value decomposition (SVD) with knowledge-aware singular values to dynamically activate knowledge based on its relevance to the task at hand. We conduct extensive experiments across a range of LLMs on tasks spanning natural language understanding (NLU), generation (NLG), instruction following, and commonsense reasoning. The experimental results demonstrate that KaSA consistently outperforms FFT and 14 popular PEFT baselines across 16 benchmarks and 4 synthetic datasets, underscoring our method's efficacy and adaptability. The source code of our method is available at \texttt{\href{https://github.com/juyongjiang/KaSA}{https://github.com/juyongjiang/KaSA}}.
\end{abstract}

\section{Introduction} \label{Introduction}
Large language models (LLMs) pretrained on massive general domain data have shown remarkable generalization ability, facilitating their application across diverse tasks \citep{zhao2023survey,touvron2023llama2,OpenAI_GPT4_2023,yoo2024hyperclova,jiang2024survey}.
The adaptation of these pretrained language models (PLMs) to specific downstream tasks generally involves full fine-tuning (FFT), where all model parameters are updated and distinct replicas of model parameters are saved for each task \citep{guo2021parameter,mao2022unipelt,gaoparameter}.
However, the increasing size of LLMs significantly raises the computational and memory costs associated with FFT, making FFT impractical in resource-constrained environments \citep{lester2021power,cai2024survey,meng2024pissa}.
Consequently, a surge of parameter-efficient fine-tuning (PEFT) methods \citep{zaken2021bitfit,li2021prefix,hu2021lora,liu2023gpt,pfeiffer2021adapterfusion,houlsby2019parameter,liu2024dora} have emerged, aiming to reduce the computational and memory costs by only updating a small set of parameters while fixing the base model \citep{mao2022unipelt, lialin2023scaling}.

Notably, LoRA \citep{hu2021lora} \revise{is} popular for its simplicity and effectiveness \citep{wang2024milora,liu2024dora,gaoparameter}.
It reparameterizes the task-specific update $\Delta \mathbf{W} \in \mathbb{R}^{n \times m}$ with a couple of low-rank matrices, $\mathbf{A}$ and $\mathbf{B}$, while keeping the base model $\mathbf{W}^{(0)} \in \mathbb{R}^{n \times m}$ unchanged during fine-tuning.
Without loss of generality, we suppose $n \ge m$ to simplify the notation.
The fine-tuning process of LoRA can be formally expressed as $\mathbf{W}^{(0)}+ \Delta \mathbf{W} = \mathbf{W}^{(0)} + \frac{\alpha}{r} \mathbf{B} \mathbf{A}^{\top}$, where $\mathbf{B} \in \mathbb{R}^{n \times r}$, $\mathbf{A} \in \mathbb{R}^{m \times r}$, $\mathbf{A}^{\top}$ is the transpose of $\mathbf{A}$, $\alpha$ is a scaling constant, and the rank $r \ll m$.
A significant advantage of LoRA is its practicality in integrating the low-rank matrices back into the base model, thereby preserving the model architecture and avoiding additional inference latency \citep{hu2021lora,han2024parameter,meng2024pissa}.

Despite LoRA's success, its initialization strategy, which employs random Gaussian noise for $\mathbf{A}$ and zeros for $\mathbf{B}$, creates an unguided subspace for the trainable parameters, causing slow convergence and suboptimal performance \citep{meng2024pissa,wang2024milora}.
To address this problem, PiSSA \citep{meng2024pissa} and MiLoRA \citep{wang2024milora} use singular value decomposition (SVD) for optimizing initialization.
SVD can decompose any matrix into three distinct matrices $(\mathbf{U}$, $\mathbf{\Sigma}$, $\mathbf{V})$, where $\mathbf{U}$ and $\mathbf{V}$ are semi-orthogonal matrices, and $\mathbf{\Sigma}$ is a diagonal matrix containing singular values sorted in descending order.
In particular, the magnitude of singular values represents the importance of parametric knowledge encapsulated in their corresponding singular vectors, with large values indicating important world knowledge and small values indicating noisy or long-tail knowledge \citep{yan2021truncated,wang2024milora,yang2023singular,sharmatruth}.
PiSSA and MiLoRA apply SVD to decompose the base model into two components: the principal components correlated with major singular values, and the residual components associated with minor singular values.
Specifically, PiSSA fine-tunes the low-rank matrices, $\mathbf{B}$ and $\mathbf{A}$, initialized with principal components, while preserving the residual components frozen, resulting in faster convergence and improved model performance \citep{meng2024pissa}. 
In contrast, MiLoRA focuses on fine-tuning $\mathbf{B}$ and $\mathbf{A}$ initialized with the minor singular value components, while fixing the principal components, aiming to boost performance and alleviate world knowledge forgetting \citep{wang2024milora}.

However, PiSSA and MiLoRA disregard two issues that can detrimentally affect model performance.
Firstly, a portion of the task-specific updates targets the weight changes of the noisy knowledge encoded in the base model, potentially leading to suboptimal performance.
Secondly, the low-rank matrices, whether initialized with the principal or residual components, inherit knowledge from the base model. These components may include information that is irrelevant to the specific downstream task, leading to conflicts within the parametric knowledge and degrading the model's representational capability.

To address these problems, we propose a PEFT method, named \textit{KaSA} (\underline{K}nowledge-\underline{a}ware \underline{S}ingular-value \underline{A}daptation), which leverages SVD with knowledge-aware singular values to dynamically activate parametric knowledge according to its relevance to downstream tasks.
Specifically, KaSA begins by performing knowledge-based SVD truncation to the base model $\mathbf{W}^{(0)}$ for removing the minor singular components $\mathbf{W}_{noise} \in \mathbb{R}^{n \times m}$ that contain noisy and long-tail knowledge \citep{gu2024sara,wang2024svd,meng2024pissa}.
This process results in an SVD-truncated model $\mathbf{W}_{world} \in \mathbb{R}^{n \times m}$ that retains essential world knowledge.
To maintain a consistent representational space between $\mathbf{W}_{world}$ and its task-specific updates $\Delta\mathbf{W}$, KaSA reparameterizes $\Delta\mathbf{W}$ in the SVD form, $\Delta\mathbf{W} = \Delta\mathbf{U} \Delta\mathbf{\Sigma} {\Delta\mathbf{V}}^{\top}$, where $\Delta\mathbf{\Sigma}$ comprises knowledge-aware singular values $(\Delta \sigma_1,...,\Delta \sigma_r)$.
The singular-value adaptation offers twofold advantages: 1) reparameterizing the task-specific updates in SVD form ensures that these updates and $\mathbf{W}_{world}$ share the same representational space, thereby preserving knowledge consistency; 2) the knowledge-aware singular values learn to activate the parametric knowledge based on its relevance to specific downstream tasks, reducing the intervention of irrelevant knowledge and enhancing model performance.

We conduct extensive experiments to fine-tune LLMs of varying sizes and architectures across a wide range of tasks, including natural language understanding (NLU), natural language generation (NLG), instruction following, and commonsense reasoning tasks.
Substantial experimental results demonstrate that our KaSA consistently outperforms FFT and \revise{14} existing popular PEFT baselines across a variety of LLMs on \revise{16 benchmarks} and 4 synthetic datasets, highlighting its efficacy and adaptability.
To summarize, in this work, our key contributions are as follows:
\begin{itemize}
    \item We propose a novel PEFT method, KaSA, which leverages SVD with knowledge-aware singular values to activate parametric knowledge based on its relevance to downstream tasks, achieving superior performance over FFT and existing popular PEFT techniques across various tasks.
    \item Our KaSA features a linear framework that allows seamless integration of the singular value adaptation module with the SVD truncated model architecture, inducing no inference latency. Furthermore, our method supports training distinct adaptation modules for different tasks, all sharing a single base model, thereby reducing the storage needs for task-switching.
    \item We conduct extensive experiments on NLU, NLG, instruction following, and commonsense reasoning tasks using popular LLMs on well-known benchmarks. Our KaSA consistently outperforms FFT and 14 PEFT baselines across different benchmarks and synthetic datasets, demonstrating its efficacy and adaptability.
    \item We make all high-quality synthetic instruction-following datasets generated by GPT4o publicly available \footnote{\href{https://huggingface.co/llama-duo}{https://huggingface.co/llama-duo}}, enabling the community to enhance the functionality of PEFT and support future research endeavors.
\end{itemize}

\section{Related Work}
\subsection{Parameter-Efficient Fine-tuning}
The increasing LLM scale presents significant challenges to efficiently adapting these models to specific tasks \citep{lialin2023scaling, zhao2023survey}.
In response, \revise{a surge of PEFT methods has emerged, reducing the computation burden by updating a minimal set of parameters during fine-tuning \citep{mao2022unipelt,karimi2021compacter,han2024parameter}. }
PEFT methods can be generally categorized into selective, additive, and re-parameterized methods \citep{ding2022delta,lialin2023scaling,xu2023parameter}.
Selective methods \citep{zaken2021bitfit,sung2021training,guo2021parameter,he2023sensitivity} train a predetermined set of the model's existing parameters while keeping the rest of the model intact.
Additive methods \citep{houlsby2019parameter,he2022towards,li2021prefix,liu2023gpt,lester2021power} introduce extra modules or parameters to fine-tune and maintain the original base model frozen.
Reparametrized methods \citep{hu2021lora,dettmers2023qlora,zhang2022adaptive,valipour2023dylora,liu2024dora} reparameterize the model's weight updates into an equivalent low-rank form for fine-tuning.
Among reparameterized approaches, LoRA stands out for its simple yet efficient mechanism of employing two low-rank matrices to approximate task-specific updates.
The fine-tuned LoRA matrices can be integrated with the base model, ensuring no inference latency.
LoRA has inspired a series of variants, each targeting specific improvements.
For instance, DyLoRA \citep{valipour2023dylora} trains the low-rank matrices across a spectrum of ranks by sorting the representation learned at different ranks during training, shortening the training time.
QLoRA \citep{dettmers2023qlora} combines 4-bit quantization with LoRA for enhanced resource efficiency. 
DoRA \citep{liu2024dora} decomposes the base model into magnitude and direction components for fine-tuning, reducing the number of trainable parameters and improving performance over LoRA.
Our method, KaSA, diverges from these reparametrized methods by employing a knowledge-aware SVD structure, enhancing the fine-tuning efficacy further.

\subsection{Singular Value Decomposition in Natural Language Processing} \label{SVD_NLP}
SVD plays a crucial role in Natural Language Processing (NLP) domain for various applications, such as model compression \citep{yuan2023asvd, wang2024svd, hsu2021language, chen2021drone}, dimensionality reduction of word embeddings \citep{tanwar2018dimensionality, shyamasundar2016twitter}, and latent semantic structure analysis \citep{deerwester1990indexing, kou2015application, horasan2019alternate}.
In the rapidly growing realm of LLMs, SVD emerges as a promising, yet relatively underexplored, technique for PEFT.
A series of SVD-based PEFT methods exploit the relationship between SVD and matrix rank to ascertain optimal ranks for specific downstream tasks.
For example, AdaLoRA \citep{zhang2022adaptive} employs SVD to reparameterize task-specific updates and adaptively determines the suitable rank through importance scoring, thus improving the model performance and parameter efficiency.
SARA \citep{gu2024sara} conducts SVD at the initialization phase to identify the appropriate rank for each layer, thereby maintaining the benefits of LoRA and boosting performance.
PiSSA \citep{meng2024pissa} and MiLoRA \citep{wang2024milora}, as mentioned in Section \ref{Introduction}, utilize SVD to optimize LoRA's initialization.
Specifically, PiSSA \citep{meng2024pissa} only fine-tunes the low-rank matrices initialized with the principal components associated with \revise{a few largest} singular values, while preserving the residual frozen. This initialization strategy facilitates faster convergence and enhanced performance. 
Conversely, MiLoRA \citep{wang2024milora} fine-tunes \revise{the minor components associated with minimal singular values}, enhancing model performance while preserving the model's world knowledge.
Unlike these methods, our method emphasizes the adaptive adjustment of singular values, allowing nuanced and dynamic activation of parametric knowledge based on its importance to downstream tasks.

\begin{figure*}[t]
	\centering
	\includegraphics[width=0.98\linewidth]{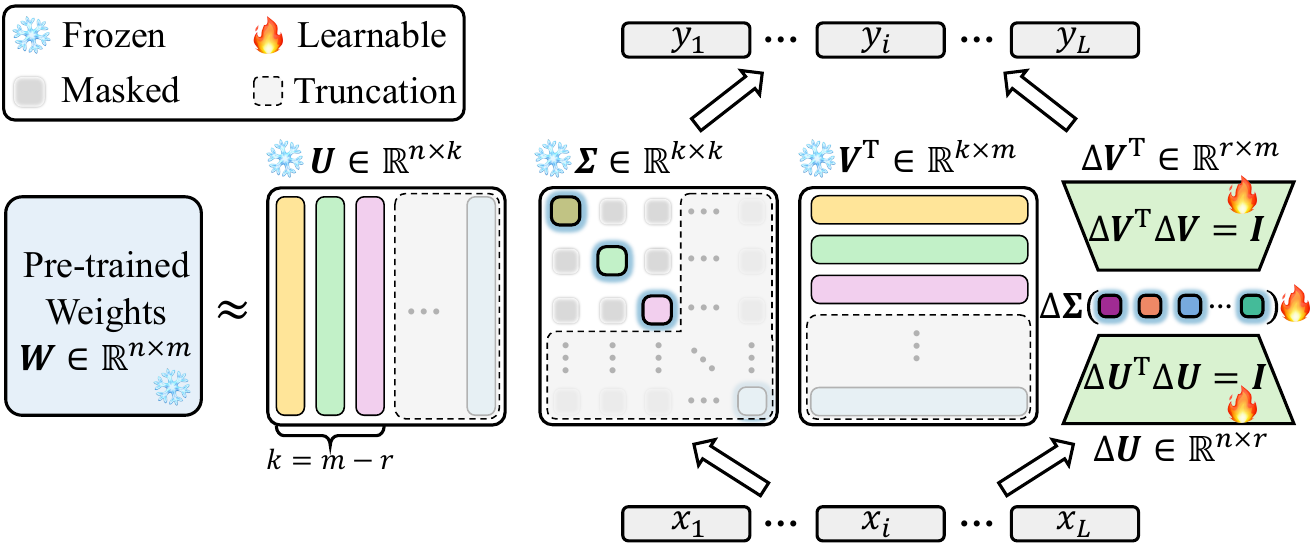}
	\caption{The architecture of our proposed KaSA encompasses two stages: (\textbf{Left}) knowledge-based SVD truncation to \revise{remove the noisy knowledge from the base model;}
    (\textbf{Right}) knowledge-aware singular-value adaptation to adjust singular values that dynamically activate knowledge across $\Delta \mathbf{W}$ model parameters based on its relevance to downstream tasks.}
\label{fig:kasa}
\vspace{-5mm}
\end{figure*}

\section{Methodology}
\revise{In this section, we commence with modeling the general PEFT process and training objective in Section \ref{sec:problem_statement}.
We subsequently provide a detailed introduction of KaSA in Section \ref{method_kasa}, followed by the description of its training objective in Section \ref{sec:training_objective}.}
\subsection{Problem Statement} \label{sec:problem_statement}
\revise{Before introducing KaSA, it is essential to delineate and model the process and objective of PEFT for LLMs based on the Transformer architecture \citep{vaswani2017attention}.
Fundamentally, PEFT is the process of training a pretrained model to a targeted task using a task-specific dataset.
It aims to minimize the divergence between the predicted probability distribution of the fine-tuned model and the actual distribution of the training data, while only modifying a small set of parameters.

Consider a pretrained model $\mathbf{W}^{(0)}$, initially parameterized by $\Theta_0$. 
To adapt this model to a particular task, we employ PEFT using a dataset $D = \{(x_l, y_l)\}_{l=1}^{Q}$ comprising $Q$ input-output instances.
The PEFT process utilizes a limited set of parameters, denoted as $\mathbf{\Psi}$, to learn the task-specific update $\triangle \Theta$, ensuring that $\vert\mathbf{\Psi}\vert\ll \vert \Theta_0\vert$.
This results in a fine-tuned model $\mathbf{W}$, parameterized by $\Theta_0 + \triangle \Theta(\mathbf{\Psi})$.
The objective is to align the predicted probability distribution of $\mathbf{W}$ with the actual distribution of training data, thereby enhancing the fine-tuned model's task performance.
The primary objective of PEFT is thus centered on the optimization of $\mathbf{\Psi}$:}
\begin{align}
\label{eq:L1}
    \mathcal{L}_1(\mathbf{\Psi}) = \sum_{(x,y)\in D}  \sum_{t=1}^{|y|} -\log({\revise{P}_{\Theta_0+\triangle \Theta(\mathbf{\Psi})}}(y_{t} | x, y_{<t}))
\end{align}

\subsection{Knowledge-Aware Singular-Value Adaptation} \label{method_kasa}
As depicted in Fig.\ref{fig:kasa}, KaSA encompasses two primary stages: 1) the knowledge-based SVD truncation, \revise{which removes the noisy knowledge from the base model}; and 2) knowledge-aware singular-value adaptation, which involves adjustment of singular values that dynamically activates parametric knowledge based on its relevance to the targeted task.

KaSA begins with a knowledge-based SVD truncation to the base model $\mathbf{W}^{(0)} \in \mathbb{R}^{n \times m}$. For simplicity of denotation, we suppose $n \ge m$. \revise{This process factories $\mathbf{W}^{(0)}$ using SVD and subsequently truncates the minor singular components $\mathbf{W}_{noise} \in \mathbb{R}^{n \times m}$, removing noisy and long-tail knowledge and resulting in a lower-rank model $\mathbf{W}_{world} \in \mathbb{R}^{n \times m}$.}
\revise{We use this refined model $\mathbf{W}_{world}$ to approximate the base model, making the adaptation of $\mathbf{W}^{(0)}$ to be resembled by that of $\mathbf{W}_{world}$:}
\begin{align}
    \mathbf{W} &= \mathbf{W}^{(0)} + \Delta \mathbf{W} = \mathbf{U}\mathbf{\Sigma}{\mathbf{V}}^{\top} + \revise{\Delta(\mathbf{U} \mathbf{\Sigma} \mathbf{V}^{\top})} = \sum_{i=1}^{m} {u_i \sigma_i {v_i}^{\top}} + \revise{\sum_{i=1}^{m} {\Delta (u_i \sigma_i v_{i}^{\top})}}\\
    &= (\mathbf{W}_{world} + \mathbf{W}_{noise}) + (\Delta \mathbf{W}_{world} + \Delta \mathbf{W}_{noise}) \\
    &= (\sum_{i=1}^{m-r} {u_i \sigma_i {v_i}^{\top}} + \sum_{i=1}^{r} {u_i \sigma_i {v_i}^{\top}}) +
    \revise{(\sum_{i=1}^{m-r} {\Delta (u_i \sigma_i v_i^{\top})} + \sum_{i=1}^{r} {\Delta (u_i \sigma_i  v_i^{\top})})}\\
    &\approx \mathbf{W}_{world} + \Delta \mathbf{W}_{world} = \sum_{i=1}^{m-r} {u_i \sigma_i {v_i}^{\top}} + \revise{\sum_{i=1}^{m-r} {\Delta (u_i \sigma_i v_i^{\top})}}
\label{truncation}
\end{align}
where $\mathbf{U} \in \mathbb{R}^{n \times m}$, $\mathbf{V} \in \mathbb{R}^{m \times m}$, and $\mathbf{V}^{\top}$ is the transpose of $\mathbf{V}$.
$\mathbf{U} = [u_1,...,u_{m}]$ and $\mathbf{V} = [v_1,...,v_{m}]$ are the corresponding left and right singular vector matrices, respectively.
The diagonal matrix $\mathbf{\Sigma} \in \mathbb{R}^{m \times m}$ contains positive singular values $(\sigma_1,...,\sigma_{m})$ sorted from high to low $({\sigma _1} \ge {\sigma _2} \ge \dotsb \geq {\sigma _ m} \ge 0)$.
The hyperparameter $r$ represents the number of \revise{truncated minor singular values}, with $r \ll m$.
The left and right singular vector matrix, $\mathbf{U}$ and $\mathbf{V}$, are semi-orthogonal:
\begin{align}
    \mathbf{U}^{\top} \mathbf{U} = \mathbf{V}^{\top} \mathbf{V} = \mathbf{I}_{m}
\label{semi_orthogonal}
\end{align}
where the identity matrix $\mathbf{I}_{m} \in \mathbb{R}^{m \times m}$.
Following the knowledge-based SVD truncation, we employ the knowledge-aware singular-value adaptation, which reparameterizes the task-specific updates of $\mathbf{W}_{world}$ in the SVD form with knowledge-aware singular values. Therefore, \revise{the weight of a model fine-tuned with KaSA} can be formally expressed as:
\begin{align}
    \mathbf{W} = \mathbf{W}^{(0)} + \Delta \mathbf{W}  
    & \approx \mathbf{W}_{world} + \eta\Delta\mathbf{U}\Delta\mathbf{\Sigma}\Delta\mathbf{V}^\top 
    = \textcolor{black}{\sum_{i=1}^{m-r} {u_i (\sigma_i) {v_i}^{\top}}} + \textcolor{black}{\eta\sum_{j=1}^{r}{\Delta u_j(\Delta \sigma_j){\Delta v_j}^{\top}}} \nonumber\\
    & \mathrm{s.t. }\quad \Delta \mathbf{U}^{\top} \Delta\mathbf{U} = \Delta \mathbf{V}^{\top} \Delta \mathbf{V} = \mathbf{I}_{r}
\end{align}
where \revise{$\mathbf{I}_{r}\in\mathbb{R}^{r \times r}$,} $\eta > 0$ is a constant scaler, the diagonal matrix $\Delta\mathbf{\Sigma} \in  \mathbb{R}^{r \times r}$ comprising \revise{learnable} knowledge-aware singular values $(\Delta\sigma_{1},...,\Delta\sigma_{r})$. \revise{The matrices $\Delta\mathbf{U}$ and $\Delta\mathbf{V}$ are semi-orthogonal, ensuring that the updates retain necessary structural properties.}

\subsection{Training Objective}
\label{sec:training_objective}
\revise{FFT typically serves as a comparative performance upper bound for PEFT methods \citep{valipour2023dylora}. Consequently, we expect that the performance of the model fine-tuned with KaSA will closely approximate that of FFT. 
We denote the FFT model as $\mathbf{W}_{fft} = \mathbf{W}^{(0)}+\Delta\mathbf{W}$.}
We impose a regularization $\Vert \mathbf{W}_{fft} - \mathbf{W}_{world} \Vert_F$, represented by the Frobenius norm, \revise{to constrain the task-specific updates.}
Based on the properties of Frobenius norms, we can further explore the boundary of the task-specific updates:
\begin{equation}
    \Vert \mathbf{W}_{fft}\Vert_F + \Vert \mathbf{W}_{world} \Vert_F \ge \Vert \mathbf{W}_{fft} - \mathbf{W}_{world} \Vert_F \ge  \Vert\Delta\mathbf{U}\Delta\mathbf{\Sigma}\Delta\mathbf{V}^\top\Vert_F = \Vert\sum_{j=1}^{r}{\Delta u_j(\Delta\sigma_j){\Delta v_j}^{\top}} \Vert_F
\label{eq:svd-loss}
\end{equation}
To stabilize the model training and extend the searching space, we introduce $\mathcal{L}_{2}$ to minimize the lower boundary of $\Vert \mathbf{W}_{fft} - \mathbf{W}_{world} \Vert_F$:
\begin{align}
    \mathcal{L}_{2}(\Delta\mathbf{\Sigma}) =  \Vert\Delta\mathbf{U}\Delta\mathbf{\Sigma}\Delta\mathbf{V}^\top\Vert_{F}^{2}
\label{l2_loss}
\end{align}
According to the Eckart–Young–Mirsky theorem \citep{eckart1936approximation}, $\mathcal{L}_{2}$ is reformulated as:
\begin{align}
     \mathcal{L}_{2}(\Delta\mathbf{\Sigma}) =  \Vert\Delta\mathbf{U}\Delta\mathbf{\Sigma}\Delta\mathbf{V}^\top\Vert_{F}^{2} = \Vert \sum_{j=1}^{r}{\Delta u_j(\Delta\sigma_j){\Delta v_j}^{\top}} \Vert_{F}^{2} = \sum_{j=1}^{r}{(\Delta\sigma_j)^2}
\label{Eckart_tbeorem}
\end{align}
Our method proposes knowledge-aware singular-value adaptation, which reparameterizes the task-specific update in the SVD form and guides $\Delta\mathbf{U}$ and $\Delta\mathbf{V}$ to conform to orthogonality.
Given this, we introduce $\mathcal{L}_{3}$ to constrain $\Delta\mathbf{U}$ and $\Delta\mathbf{V}$ adhere to orthogonality, such that:
\begin{equation}
    \mathcal{L}_{3}(\mathbf{\Psi}) = {\left\Vert \Delta\mathbf{U}^{\top} \Delta\mathbf{U} - \revise{\mathbf{I}_r}\right\Vert}_F + {\left\Vert \Delta\mathbf{V}^{\top} \Delta\mathbf{V} - \revise{\mathbf{I}_r} \right\Vert}_F
\end{equation}
Overall, our methods leverage $\mathcal{L}_1$, $\mathcal{L}_2$, and $\mathcal{L}_3$ to serve jointly for optimizing the model's task performance while adhering to SVD structure. For adjusting $\mathcal{L}_2$ and $\mathcal{L}_3$, we introduce $\beta > 0$ and $\gamma >0$ as their corresponding scalers. The overall training objective of KaSA can be expressed as:
\begin{equation}
    \mathcal{L}(\mathbf{\Psi},\Delta\mathbf{\Sigma}) = \min_{\mathbf{\Psi},\Delta\mathbf{\Sigma}} (\mathcal{L}_1(\mathbf{\Psi},\Delta\mathbf{\Sigma}) + \beta \mathcal{L}_2(\Delta\mathbf{\Sigma}) + \gamma \mathcal{L}_3(\mathbf{\Psi}))
\end{equation}
\revise{We present the PyTorch-style pseudocode for KaSA and its training objective in Appendix \ref{sec:pseudocode}.}

\section{Experiments}
In this section, we evaluate KaSA's efficacy across different downstream tasks, including natural language understanding (NLU), natural language generation (NLG) (see Appendix \ref{sec:addition_nlg}), instruction following, and commonsense reasoning.
For NLU tasks, we evaluate KaSA with RoBERTa \citep{liu2021robustly} and DeBERTaV3 \citep{he2022debertav3} on the GLUE \citep{wang2018glue} benchmark. 
For NLG tasks, we assess our method with GPT-2 \citep{radford2019language} on the E2E NLG Challenge \citep{novikova2017e2e} benchmark.
We further assess instruction following performance with popular LLMs, including LLaMA3 8B \citep{llama3}, Mistal 7B \citep{jiang2023mistral}, Gemma 7B \citep{team2024gemma}, \revise{and LLaMA2 13B \citep{touvron2023llama2}}.
These models are fine-tuned with different PEFT methods using four synthetic datasets generated by GPT4o, each tailored to summarization, classification, coding, and closed QA.
GPT4o is then employed as a judge to evaluate the fine-tuned models' performance, assigning scores on a scale of 10.
\revise{We also follow \citep{kopiczko2023vera} and \citep{gaoparameter} to fine-tune the four models on the Alpaca dataset \citep{taori2023stanford} and report evaluation results on MT-Bench, with GPT4 serving as the judge, yielding scores within 10.}
Additionally, we substantiate KaSA's generality by fine-tuning LLaMA2 7B and LLaMA3 8B models on the Commonsense170K dataset \citep{hu2023llm}, which includes training sets from eight commonsense reasoning datasets, and evaluating them on individual test sets of these constituent datasets.
Finally, we conduct ablation studies to investigate the impacts of different components, budget parameter scalability, and the distribution of knowledge-aware singular values across various layers.
All experiments are conducted on NVIDIA A100-SXM4 (80GB) GPUs, except for the NLU experiments, which are conducted on NVIDIA GeForce RTX 3090 (24GB) GPUs.

\subsection{Baselines}
\revise{We compare KaSA with FFT and 14 PEFT baselines to substantiate its efficacy and robustness: \\
$ \bullet $ \textbf{Adapter-based methods} We consider four representative Adapter tuning methods as baselines: 1) $\text{Adapter}^{\text{H}}$ \citep{houlsby2019parameter}; 2) $\text{Adapter}^{\text{D}}$ \citep{ruckle2021adapterdrop}; 3) $\text{Adapter}^{\text{L}}$ \citep{lin2020exploring}; and 4) $\text{Adapter}^{\text{P}}$ \citep{pfeiffer2021adapterfusion}. \\
$ \bullet $ \textbf{LoRA-based methods} We select LoRA and its variants: 1) LoRA \citep{hu2021lora}; 2) DyLoRA \citep{valipour2023dylora}; 3) VeRA \citep{kopiczko2023vera}; and 4) DoRA \citep{liu2024dora}. \\
$ \bullet $ \textbf{SVD-based methods} Considering that our method is associated with SVD, we chose SVD-based PEFT baselines: 1) AdaLoRA \citep{zhang2022adaptive}; 2) PiSSA \citep{meng2024pissa}; 3) MiLoRA \citep{wang2024milora}; 4) SARA \citep{gu2024sara}; and 5) CorDA \citep{yangcorda}. \\
$ \bullet $ \textbf{Other methods} Apart from the aforementioned baselines, we also consider other important fine-tuning methods: 1) FFT; and 2) BitFit \citep{zaken2021bitfit}. \\
To ensure a fair comparison with these baselines, we meticulously replicate the experimental configurations as described in previous studies \citep{hu2021lora,zhang2022adaptive,gu2024sara}.
Introductions of the baselines and comprehensive details of the experimental setup are provided in Appendix \ref{sec:peft_baselines} and Appendix \ref{sec:train_details}, respectively. \\}
\begin{table}[t]
\centering
\caption{Performance of RoBERTa-base ($\text{RoB}_\text{base}$) and RoBERTa-large ($\text{RoB}_\text{large}$) with different adaptation methods on 6 datasets of the GLUE benchmark. 
We report the overall (matched and mismatched) accuracy for MNLI, Matthew’s correlation coefficient (Mcc.) for CoLA, Pearson correlation coefficient (Pcc.) for STS-B, and accuracy (Acc.) for all the remaining tasks. 
We report the average result of five runs with different random seeds.
The best results for each dataset are shown in \textbf{bold}.
Higher is better for all metrics.
}
\label{tab:glue_datasets}
\resizebox{0.85\linewidth}{!}{
\begin{tabular}{l|r|ccccccc}
\toprule
\multirow{2}*{\textbf{ Model(Method)}} & \textbf{\multirow{2}*{\makecell[r]{\# Trainable \\Parameters}}} & {\textbf{ SST-2}} & {\textbf{ MRPC}}  & {\textbf{ CoLA}} & {\textbf{ QNLI}} & {\textbf{ RTE}} & {\textbf{ STS-B}} & {\textbf{ All}} \\
~ & ~ & {(Acc.)} & {(Acc.)} &  {(Mcc.)} & {(Acc.)} & {(Acc.)} & {(Pcc.)} & {Avg.} \\
\midrule 
$\text{RoB}_\text{base}$(FFT) & 125.0M  & 94.8     & 90.2         & 63.6     & 92.8    & 78.7   & 91.2   & 85.2     \\ 
$\text{RoB}_\text{base}$(BitFit) &   0.1M  & 93.7     & \textbf{92.7}         & 62.0     & 91.8      & {81.5}   & 90.8   &  85.4    \\ 
$\text{RoB}_\text{base}$($\text{Adpt}^{\text{D}}$)  &   0.3M  & 94.2     & 88.5         & 60.8     & 93.1        & 71.5   & 89.7   & 83.0     \\ 
$\text{RoB}_\text{base}$($\text{Adpt}^{\text{D}}$) &   0.9M  & 94.7     & 88.4         & 62.6     & 93.0       & 75.9   & 90.3   & 84.2     \\ 
$\text{RoB}_\text{base}$(LoRA) &   0.3M & 95.1     & 89.7         & 63.4     & \textbf{93.3}        & 78.4   & \textbf{91.5}   & 85.2     \\ 
$\text{RoB}_\text{base}$(AdaLoRA) & 0.3M & 94.5 & 88.7 & 62.0 & 93.1 & 81.0 & 90.5 & 85.0 \\
$\text{RoB}_\text{base}$(DyLoRA) & 0.3M &  94.3 & 89.5 & 61.1 & 92.2 & 78.7 & 91.1 & 84.5 \\
$\text{RoB}_\text{base}$(PiSSA) & 0.3M & 95.0 &  88.2 &  65.5 &  92.0 &  75.1 & 90.4 & 84.4 \\ 
$\text{RoB}_\text{base}$(MiLoRA)  & 0.3M  & 94.6 & 88.7 & 63.1 & 92.8 & 80.5 & {91.3}  & 85.2 \\
$\text{RoB}_\text{base}$($\text{KaSA})$  & 0.3M & \textcolor{black}{\textbf{95.2}} & \textcolor{black}{{90.7}} & \textcolor{black}{\textbf{65.8}} & \textcolor{black}{\textbf{93.3}} & \textcolor{black}{\textbf{81.6}} & \textcolor{black}{91.1} & \textbf{86.3}\\
\midrule
$\text{RoB}_\text{large}$(FFT) &   355.0M  & 96.4    & 90.9         & 68.0    & 94.7      & 86.6   & 92.4   & 88.2     \\ 
$\text{RoB}_\text{large}$($\text{Adpt}^{\text{P}}$) &   3.0M  & 96.1     & 90.2         & 68.3     & {94.8}    & 83.8   & 92.1   & 87.6    \\ 
$\text{RoB}_\text{large}$($\text{Adpt}^{\text{P}}$) &   0.8M  & 96.6     & 89.7         & 67.8     & {94.8}      & 80.1   & 91.9   & 86.8     \\ 
$\text{RoB}_\text{large}$($\text{Adpt}^{\text{H}}$) &   6.0M  & 96.2     & 88.7         & 66.5     & 94.7      & 83.4   & 91.0   & 86.8     \\ 
$\text{RoB}_\text{large}$($\text{Adpt}^{\text{H}}$) &   0.8M  & 96.3     & 87.7         & 66.3     & 94.7       & 72.9   & 91.5   & 84.9    \\ 
$\text{RoB}_\text{large}$(LoRA) &   0.8M & 96.2     & 90.2         & 68.2     & {94.8}      & 85.2   & {92.3}   & 87.8     \\ 
$\text{RoB}_\text{large}$($\text{KaSA}$)  & 0.8M & \textcolor{black}{\textbf{96.9}} & \textcolor{black}{\textbf{91.2}} &  \textcolor{black}{\textbf{69.4}} & \textcolor{black}{\textbf{94.9}}  & \textcolor{black}{\textbf{88.8}} & \textcolor{black}{\textbf{92.5}} & \textbf{89.0}\\
\bottomrule 
\end{tabular}
}
\end{table}

\subsection{Natural Language Understanding}\label{sec:nlu}
\textbf{Models and Datasets.} 
For NLU tasks, our method involves fine-tuning foundation models such as RoBERTa-base (125M), RoBERTa-large (355M) \citep{liu2021robustly}, and DeBERTaV3-base \citep{he2022debertav3} using the GLUE (General Language Understanding Evaluation) benchmark \citep{wang2018glue}. 
The GLUE benchmark encompasses a wide array of datasets designed to test various aspects of NLU, including question answering, natural language inference, sentiment analysis, and textual entailment.
In this context, our evaluation is conducted \revise{across 6 datasets from the GLUE: SST-2, MRPC, CoLA, QNLI, RTE, and STS-B.}
Detailed statistical information about the GLUE benchmark can be found in Appendix \ref{sec:glue_benchmark}.

\textbf{Implementation Details.}
Basically, we follow the experimental setup applied in \citep{hu2021lora,zhang2022adaptive} to ensure a fair comparison. 
We randomly initialize the knowledge-aware singular values without bias, which only introduces negligible $r$ coefficients in each layer. 
For all evaluated datasets in GLUE, we meticulously tune the hyperparameters, including the learning rates $lr \in [\text{1E-5}, \text{1E-3}]$, the rank of SVD truncation $k \in \{1, 2, 4, 8, 16, 32, 64, 128\}$, and two trade-off loss coefficients $\beta \in [\text{1E-5}, 1]$ and $\gamma \in [\text{1E-5}, 1]$. 
The results we present are the median outcomes from 5 runs, each conducted with a distinct random seed. 
To maintain fair trainable parameters, we fine-tune the \texttt{query} and \texttt{value} weights in each Transformer block and set a rank $r = 8$ across all datasets. 
More detailed hyperparameters are presented in Appendix \ref{sec:details_nlu}.

\textbf{Main Results.}
Table \ref{tab:glue_datasets} presents the performance of RoBERTa-base and RoBERTa-large models fine-tuned using our KaSA in contrast to PEFT baselines.
KaSA achieves the best performance across all datasets except MRPC and STS-B for the RoBERTa-base model. 
Notably, KaSA registers the highest average performances for both RoBERTa models: 86.3\% for RoBERTa-base and 89.0\% for RoBERTa-large. 
This underscores the effectiveness, adaptability, and scalability of our proposed approach.
In a significant comparison with FFT, our KaSA, which utilizes merely up to 0.24\% (approximately 0.3M/125.0M) of trainable parameters, outperforms FFT in 13 out of 14 scenarios and matches its performance on the STS-B dataset for the RoBERTa-base model.
The results from DeBERTaV3-base are presented in Appendix \ref{sec:addition_nlu}.

\subsection{Instruction Following}
\textbf{Models and Datasets.}
To validate KaSA's adaptability and versatility, we extend our experiments to include instruction tuning of LLaMA3 8B \citep{llama3}, Mistral 7B \citep{jiang2023mistral}, Gemma 7B \citep{team2024gemma}, and \revise{LLaMA2 13B \citep{touvron2023llama2}}.
We fine-tune the models using four synthetic instruction-following datasets produced by GPT4o, each containing 128K samples, covering tasks such as summarization, classification, coding, and closed QA.
\revise{Additionally, we fine-tune using the Alpaca dataset \citep{taori2023stanford} and report the evaluation results on MT-Bench \citep{zheng2023judging}, with GPT4 serving as the judge, yielding scores within 10.}
\revise{The detailed processing and statistical information of the synthetic datasets, Alpaca, and MT-Bench are presented in Appendix \ref{sec:synthetic_dataset} and \ref{sec:alpaca_MTBench}, respectively.}


\begin{table}[t]
\centering
\caption{\revise{Instruction following evaluation results with average scores for the most popular LLMs fine-tuned on the 128k synthetic datasets and the Alpaca dataset, and evaluated by GPT4o and GPT4 with the scores within 10 on test subsets and MT-Bench, respectively. 
}} 
\label{tab:instruction_tuning}
\resizebox{0.9\textwidth}{!}{
\begin{tabular}{l|l|r|cccc|c}
\toprule
\textbf{Model} & \textbf{Method} & \textbf{\makecell[r]{\# Trainable \\Parameters}} & \textbf{Classification} & \textbf{Summarization} & \textbf{Coding} & \textbf{Closed QA} & \textbf{MT-Bench}  \\ 
\midrule
\multirow{6}{*}{Gemma 7B} 
& w/o FT & - & 2.41 & 2.28 & 3.07 & 2.95 & 2.56 \\
& FFT  & 8.54B & 5.58   & 7.78 & 7.61 & \textbf{8.88} & 4.69 \\
& LoRA & 3.21M & 5.98  & 7.29  & 7.75  & 8.18   & 4.32  \\
& PiSSA & 3.21M & 6.23   & 7.88  & 7.80  & 8.22  & 4.66  \\
& MiLoRA & 3.21M & 6.30  & 7.62  & 7.71  & 8.27  & 4.53  \\
& \textbf{KaSA} & 3.22M & \textbf{6.88} & \textbf{7.92} & \textbf{8.01} & 8.69 & \textbf{4.97} \\
\midrule
\multirow{6}{*}{Mistral 7B} 
& w/o FT & - & 2.31 & 2.81 & 2.32 & 3.02 & 1.16 \\
& FFT  & 7.25B & \textbf{6.73} & \textbf{7.18} & \textbf{7.53} & \textbf{8.75} & 4.22 \\
& LoRA & 3.40M & 5.07  & 5.72  & 6.17  & 7.39  & 4.18  \\
& PiSSA & 3.40M & 5.46  & 5.86 & 6.41  & 7.24  & 4.24  \\
& MiLoRA & 3.40M & 5.33  & 5.89 & 6.52  & 7.28  & 4.29  \\
& \textbf{KaSA} & 3.41M & 5.72 & 6.82 & 6.74 & 7.75 & \textbf{4.58} \\
\midrule
\multirow{6}{*}{LLaMA3 8B} 
& w/o FT & - & 2.04 & 2.03 & 2.86 & 3.33 & 3.11 \\
& FFT  & 8.03B & 5.44 & 7.80 & 7.59 & \textbf{8.90} & 4.11 \\
& LoRA & 3.40M & 6.12  & 7.20  & 7.37  & 6.02  & 4.19  \\
& PiSSA & 3.40M & 6.35  & 7.31  & 7.59  & 6.18  & 4.26  \\
& MiLoRA & 3.40M & 6.37  & 7.61  & 7.65  & 6.39  & 4.32  \\
& \textbf{KaSA} & 3.41M & \textbf{6.55} & \textbf{7.83} & \textbf{7.89} & 6.81 & \textbf{4.71} \\
\midrule
\multirow{6}{*}{LLaMA2 13B} 
& w/o FT & - & 1.00 & 1.08 & 1.01 & 1.27 & 1.01 \\
& FFT  & 13.02B & 5.86         & \textbf{7.93} & 7.88 & \textbf{8.97} & 4.37 \\
& LoRA & 6.55M & 6.23          & 7.38  & 7.54  & 6.25  & 4.43 \\
& PiSSA & 6.55M & 6.47         & 7.45  & 7.83  & 6.54  & 4.39  \\
& MiLoRA & 6.55M & 6.45        & 7.63  & 7.85  & 6.82  & 4.51  \\
& \textbf{KaSA} & 6.56M & \textbf{6.86} & 7.92 & \textbf{8.09} & 7.12 & \textbf{4.95} \\
\bottomrule
\end{tabular}
}
\vspace{-1mm}
\end{table}

\textbf{Implementation Details.}
Following the experimental setup in \citep{park2024llamaduo}, we use the summarization, classification, coding, and closed QA subsets from the ``No Robots'' \citep{no_robots} dataset as seeds to create distinct synthetic datasets via GPT4o.
We fine-tune the mentioned LLMs using these datasets and then prompt each fine-tuned model to generate four responses based on prompts sampled from the test subsets of the seed dataset. 
To ensure fair comparisons, we maintain a consistent fine-tuning and inference configuration across all fine-tuned models. 
We subsequently use GPT4o as a judge to apply single-answer grading strategies to evaluate the response quality of the fine-tuned LLMs on a scale from \revise{1 to 10}. 
\revise{For the Alpaca dataset, we fine-tune the specified models and prompt them to generate responses to questions from MT-Bench, with GPT4 serving as a judge, assigning scores within 10.}
Detailed prompts for data synthesis and performance evaluation, along with hyperparameter settings, are presented in Appendix \ref{sec:synthetic_dataset}, \ref{sec:prompt_templates}, and \ref{sec:instruction_following_config}, respectively.

\textbf{Main Results.}
\revise{In Table \ref{tab:instruction_tuning}, the results show that KaSA consistently surpasses LoRA, PiSSA, and MiLoRA across four 128k synthetic datasets, regardless of the model used.
Notably, Gemma 7B and LLaMA3 8B, fine-tuned with KaSA, even surpass FFT in the classification, summarization, and coding datasets.
In the evaluation using MT-Bench, KaSA consistently outperforms FFT and PEFT baselines on all models, showing remarkable efficacy.
With significance tests showing ($p < 0.05$) in 9 out of 12 experimental settings on MT-Bench, KaSA demonstrates significant performance improvements over LoRA, PiSSA, and MiLoRA.
These results further highlight the effectiveness, robustness, and adaptability of our method.}

\subsection{Commonsense Reasoning}
\textbf{Models and Datasets.}
Following \citep{wang2024milora}, we fine-tune the LLaMA2 7B \citep{touvron2023llama} and the LLaMA3 8B \citep{llama3} models using the Commonsense170K dataset, aiming to conduct a comprehensive evaluation across eight well-known commonsense reasoning tasks: BoolQ \citep{clark2019boolq}, PIQA \citep{bisk2020piqa}, SIQA \citep{sap2019socialiqa}, HellaSwag \citep{zellers2019hellaswag}, WinoGrande \citep{sakaguchi2021winogrande}, ARC-e, ARC-c \citep{clark2018think}, and OBQA \citep{mihaylov2018can}.

\textbf{Implementation Details.}
To ensure a fair comparison, we implement our KaSA within the LLM-Adapters framework \footnote{\href{https://github.com/AGI-Edgerunners/LLM-Adapters}{https://github.com/AGI-Edgerunners/LLM-Adapters}} \citep{hu2023llm}, following MiLoRA \citep{wang2024milora}. We adhere strictly to the hyperparameter configurations for training and evaluation as specified by \citep{wang2024milora} and \citep{hu2023llm}, \textbf{without any tuning}, such as tuning the training epochs and learning rate.
For detailed hyperparameters utilized, refer to Appendix \ref{sec:commonsense_reasoning}.

\begin{table}[t] 
\centering 
\caption{Performance comparison of LLaMA2 7B and LLaMA3 8B with different adaptation methods on eight 
commonsense reasoning datasets. The symbol $\dag$ indicates that the results are taken from \citep{wang2024milora}. The best results are shown in \textbf{bold}. Higher is better for all tasks. $*$ denotes that the best results do not surpass ChatGPT.
}
\resizebox{\textwidth}{!}{
\begin{tabular}{ll|ccccccccc}
\toprule
\textbf{Model} & \textbf{Method} & \textbf{BoolQ} & \textbf{PIQA} & \textbf{SIQA} & \textbf{HellaSwag}& \textbf{WinoGrande} & \textbf{ARC-e} & \textbf{ARC-c} & \textbf{OBQA} & \textbf{Avg.} \\ 
\midrule
ChatGPT$\dag$ & - & 73.1 & 85.4 & 68.5  & 78.5 & 66.1 & 89.8 & 79.9 & 74.8 & 77.0 \\
\midrule
\multirow{4}*{LLaMA2 7B} & LoRA$\dag$ & 69.8 & 79.9  & 79.5  & 83.6  & 82.6  & 79.8  & 64.7  & 81.0 & 77.6  \\
& PiSSA$\dag$ & 67.6  & 78.1  & 78.4  & 76.6  & 78.0  & 75.8  & 60.2  & 75.6  & 73.8   \\
& MiLoRA$\dag$ & 67.6  & 83.8  & 80.1  & 88.2  & 82.0  & 82.8  & 68.8  & 80.6  & 79.2  \\
& KaSA & \textbf{73.6} & \textbf{84.4}$^*$ & \textbf{80.2} & \textbf{91.5} & \textbf{84.5} & \textbf{84.7}$^*$ & \textbf{72.1}$^*$ & \textbf{81.2} & \textbf{81.5}   \\
\midrule
\multirow{4}*{LLaMA3 8B} & LoRA$\dag$ & 70.8 & 85.2 & 79.9 & 91.7 & 84.3 & 84.2 & 71.2 & 79.0 & 80.8 \\
& PiSSA$\dag$ & 67.1 & 81.1 & 77.2 & 83.6 & 78.9 & 77.7 & 63.2 & 74.6 & 75.4 \\
& MiLoRA$\dag$ & 68.8 & 86.7 & 77.2 & 92.9 & \textbf{85.6} & 86.8 & 75.5 & 81.8 & 81.9 \\
& KaSA & \textbf{73.6} & \textbf{88.1} & \textbf{80.4} & \textbf{94.7} & 85.5 & \textbf{89.7}$^*$ & \textbf{79.4}$^*$ & \textbf{85.6} & \textbf{84.6} \\
\bottomrule
\end{tabular}
}
\label{tab:commonsense_reasoning}
\end{table}

\textbf{Main Results.}
As illustrated in Table \ref{tab:commonsense_reasoning}, KaSA consistently surpasses all established baselines for both LLaMA2 7B and LLaMA3 8B across all eight benchmarks when using identical hyperparameter settings. 
Notably, KaSA achieves the highest average score, reflecting significant performance improvements across a diverse range of reasoning tasks. 
These results, obtained from rigorously controlled comparisons, align with our observations in NLU, NLG, and instruction following tasks. This consistency further corroborates the robustness and superiority of our method.

\begin{figure*}[t]
	\centering
	\includegraphics[width=0.95\linewidth]{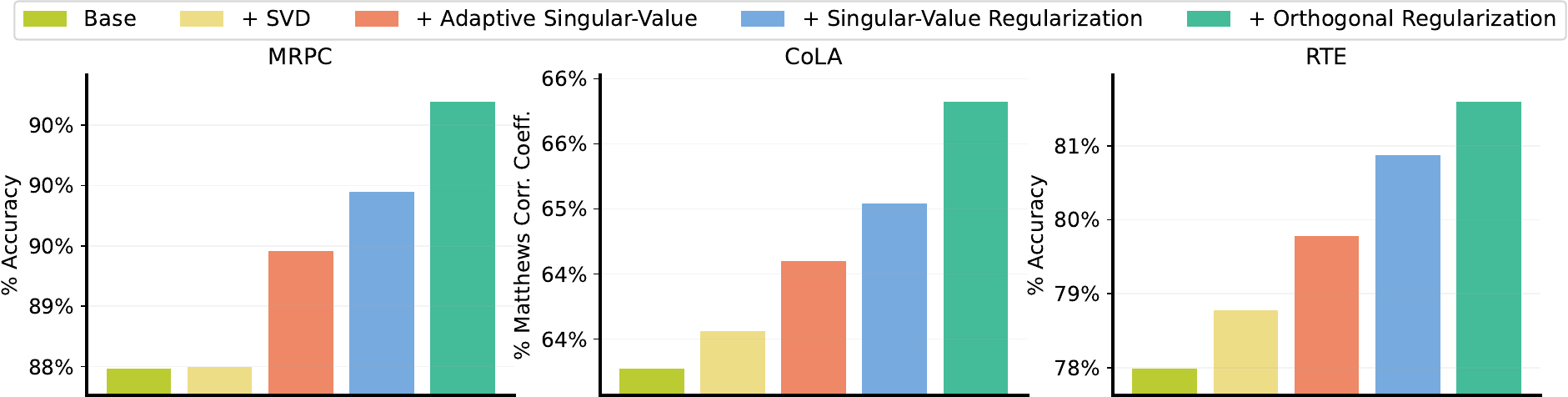}
	\caption{Components ablation study about knowledge-based SVD truncation, knowledge-aware singular value adaptation, singular value regularization $\mathcal{L}_{2}$, and orthogonal regularization $\mathcal{L}_{3}$ on MRPC, CoLA, and RTE datasets.
 }
\label{fig:ablation}
\vspace{-10pt}
\end{figure*}

\begin{figure*}[t]
	\centering
	\includegraphics[width=0.95\linewidth]{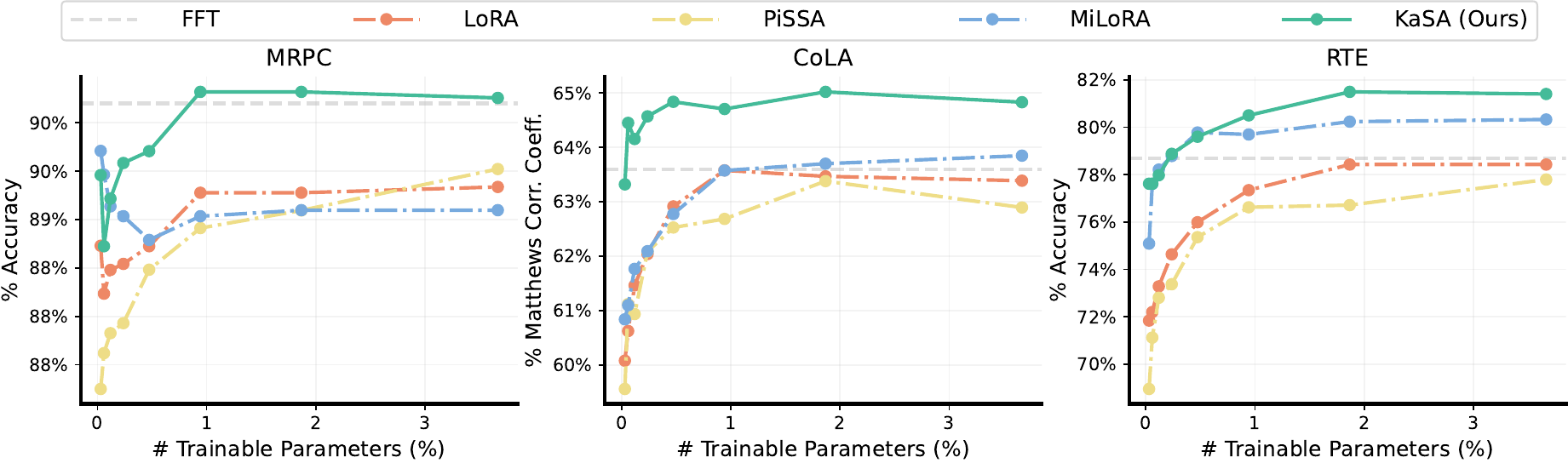} 
	\caption{Budget parameter scalability of fine-tuning RoBERTa-base with LoRA, PiSSA, MiLoRA, and KaSA on MRPC, CoLA, and RTE datasets.}
\label{fig:budget}
\vspace{-10pt}
\end{figure*}

\subsection{In-depth Analysis} \label{In-depth}
\textbf{Components Ablation Study.}
Our method encompasses four principle components: knowledge-based SVD truncation, knowledge-aware singular value adaptation, singular value regularization $\mathcal{L}_{2}$, and orthogonal regularization $\mathcal{L}_{3}$.
To examine the collective contributions of these components, we conduct ablation experiments on MRPC, CoLA, and RTE datasets from GLUE \revise{using the RoBERTa-base}.
Specifically, we compare KaSA with the following variants: 
(1) standard LoRA (as the base); 
(2) SVD truncation $+$ LoRA; 
(3) SVD truncation $+$ knowledge-aware singular-value adaptation; 
(4) SVD truncation $+$ knowledge-aware singular-value adaptation $+$ $\mathcal{L}_{2}$; 
(5) SVD truncation $+$ knowledge-aware singular-value adaptation $+$ $\mathcal{L}_{2}$ $+$ $\mathcal{L}_{3}$.
From the results in Figure \ref{fig:ablation}, we observe that the model performances continually increase as more components are involved in the fine-tuning. 
The fifth bar in Figure \ref{fig:ablation} shows that variant (5), the full implementation of KaSA, achieves significant performance improvements across all three datasets.
Conversely, excluding any of these components results in performance declines ranging from 2.05\% to 3.25\%, underscoring their collective importance in enhancing KaSA's effectiveness.
\revise{Additional results of the components ablation study on SST-2, QNLI, and STS-B datasets are detailed in Appendix \ref{sec:additional_ablation}.}

\textbf{Budget Parameter Scalability.}
We compare the performance of fine-tuning RoBERTa-base with \revise{LoRA, PiSSA, MiLoRA, and KaSA} across various scales of trainable parameters.
Specifically, we employ \revise{these methods} to the \texttt{query} and \texttt{value} weights of the transformer block and use a range of ranks $r = \{1,2,4,8,16,32,64,128\}$ to control the parameter scales.
Figure \ref{fig:budget} shows that KaSA consistently outperforms LoRA, \revise{as well as the SVD-based baselines,} at equivalent parameter scales across various datasets, indicating our method's efficacy and robustness.
Moreover, we observe that enlarging trainable parameter scales does not invariably result in performance improvement.
Notably, both methods peak in performance at $r = 8$, with KaSA enhancing LoRA by 1.96\% on MRPC, 2.05\% Mcc. on CoLA, and 2.53\% Acc. on RTE.

\begin{figure*}[t]
\centering
    \includegraphics[width=\linewidth]{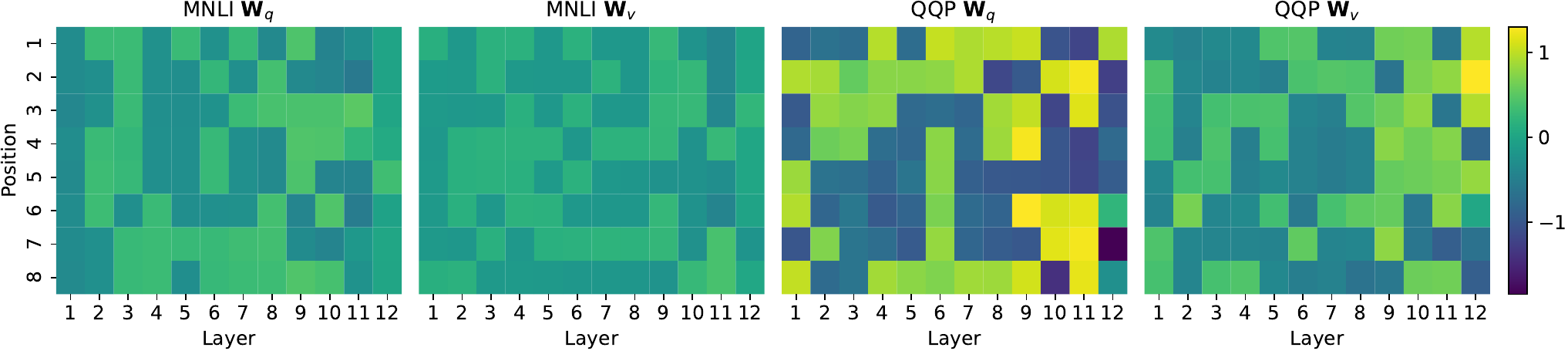}
    \vspace{-12pt}
    \caption{The final distribution of knowledge-aware singular values for $\mathbf{W}_q$ and $\mathbf{W}_v$ upon fine-tuning the RoBERTa-base model on the MNLI and QQP benchmarks.
    In this context, the $x$-axis corresponds to the layer index, and the $y$-axis denotes the position index. Each value signifies the relevance of the associated knowledge.}
\label{fig:svd_diag}
\vspace{-9pt}
\end{figure*}

\textbf{Knowledge-Aware Singular-Value.}
The conventional FFT, which updates all parameters indiscriminately, often incorporates irrelevant or minimally contributory knowledge to the task at hand, leading to overfitting and a decline in model generalization capability \citep{valipour2023dylora}. 
To this end, we propose a novel knowledge-aware singular value module to adaptively activate the relevant task-specific knowledge. 
To validate our motivation, we visualize the knowledge-aware singular values of $\mathbf{W}_q$ and $\mathbf{W}_v$ when fine-tuning RoBERTa-base on the MNLI and QQP benchmarks, as depicted in Figure \ref{fig:svd_diag}. 
We can clearly observe that different scales of singular values are allocated across different layers, indicating that it dynamically prioritizes knowledge across parameters. 
\vspace{-5pt}

\section{Conclusion}
In this paper, we introduce a PEFT method, KaSA, which incorporates SVD with knowledge-aware singular values for dynamic activation of parametric knowledge according to their relevance to the given tasks.
\revise{KaSA commences with knowledge-based SVD truncation of minor singular value components to remove noisy knowledge within the base model. 
Subsequently, it reparameterizes task-specific updates in the SVD form, leveraging knowledge-aware singular values for dynamic knowledge activation according to relevance.
Our extensive experiments on various LLMs across tasks in NLU, NLG, instruction following, and commonsense reasoning reveal that KaSA consistently surpasses FFT and a variety of prevailing PEFT baselines across well-known benchmarks and our synthetic datasets, highlighting the superiority of our method.}

\section*{Acknowledgments}
Jing Tang's work was partially supported by National Key R\&D Program of China under Grant No.\ 2023YFF0725100 and No.\ 2024YFA1012701, by the National Natural Science Foundation of China (NSFC) under Grant No.\ 62402410 and No.\ U22B2060, by Guangdong Provincial Project (No.\ 2023QN10X025), by Guangdong Basic and Applied Basic Research Foundation under Grant No.\ 2023A1515110131, by Guangzhou Municipal Science and Technology Bureau under Grant No.\ 2023A03J0667 and No.\ 2024A04J4454, by Guangzhou Municipal Education Bureau (No.\ 2024312263), and by Guangzhou Municipality Big Data Intelligence Key Lab (No.\ 2023A03J0012), Guangzhou Industrial Information and Intelligent Key Laboratory Project (No.\ 2024A03J0628) and Guangzhou Municipal Key Laboratory of Financial Technology Cutting-Edge Research (No.\ 2024A03J0630).
This work was also supported by IITP grant funded by the Korea government(MSIT)[RS-2023-00215959, Development of Access Agnostic wired and wireless integrated optical access technology].

\bibliography{ref}
\bibliographystyle{iclr2025_conference}

\clearpage
\appendix
\section{Pseudocode for KaSA} \label{sec:pseudocode}
\begin{algorithm}[h]
   \caption{PyTorch-style pseudocode for \textbf{KaSA}.}
   \label{algo:KaSA}
    \definecolor{codeblue}{rgb}{0.25,0.5,0.5}
    \definecolor{black}{rgb}{0.0, 0.0, 0.0}
    \lstset{
      basicstyle=\color{black}\fontsize{7.2pt}{7.2pt}\ttfamily\bfseries,
      commentstyle=\fontsize{7.2pt}{7.2pt}\color{codeblue},
      keywordstyle=\fontsize{7.2pt}{7.2pt}\color{black},
      stringstyle=\color{black},
      identifierstyle=\color{black},
      frame=none
    }
\begin{lstlisting}[language=python]
class KaSA(nn.Module):
    def __init__(self,
	rank: int = 8, # kasa rank
	alpha: int = 16, # kasa alpha
	base_layer: nn.Module # pre-trained layer
    ):
        # definitions
        self.r = rank
        self.alpha = alpha
        self.scaling = alpha / rank
        self.in_features, self.out_features = base_layer.in_features, base_layer.out_features

        # Step 1: knowledge-based SVD truncation
        self.svd_rank = self.in_features - self.r
        U, S, Vh = torch.linalg.svd(base_layer.weight.data, full_matrices=False)
        base_layer.weight.data = U[:, :self.svd_rank] @ torch.diag(S[:self.svd_rank]) @ Vh[:self.svd_rank, :]
        self.base_layer = base_layer

        # Step 2: knowledge-aware singular-value adaptation
        self.delta_v = nn.Linear(self.in_features, self.r, bias=False)
        self.delta_sigma = nn.Parameter(torch.randn(self.r), requires_grad=True)
        self.delta_u = nn.Linear(self.r, self.out_features, bias=False)

    def forward(self, x: torch.Tensor):
        # Step 3: merge W + Delta_W (Eq.7)
        Delta_W = self.delta_u @ torch.diag(self.delta_sigma) @ self.delta_v
        result = self.base_layer(x)
        result = result + torch.einsum('ijk,kl->ijl', x, Delta_W) * self.scaling
        return result
 
def regularization_loss(
    model: nn.Module,
    beta: float, 
    gamma: float
):
    # definitions
    l2_loss = 0.0 
    l3_loss = 0.0 
    num_param = 0
    for name, param in model.named_parameters():
        if param.requires_grad:
            # singular value regularization
            if 'delta_sigma' in name:
                num_param += 1
                diag_norm = torch.sum(param ** 2)
                l2_loss += diag_norm
            # orthogonal regularization
            elif 'delta_v' in name or 'delta_u' in name:
                if 'delta_v' in name:
                    matmul_result = torch.matmul(param.T, param)
                else:  
                    matmul_result = torch.matmul(param, param.T)

                I = torch.eye(matmul_result.size(0), device=matmul_result.device)
                diff_I = matmul_result - I
                matrix_loss = torch.norm(diff_I, p='fro')
                l3_loss += matrix_loss
    auxi_loss = (beta * l2_loss + gamma * l3_loss) / num_param if num_param > 0 else 0.0
    return auxi_loss
\end{lstlisting}
\end{algorithm}

\section{Baselines} \label{sec:peft_baselines}
To demonstrate its efficacy and robustness, we evaluate KaSA against FFT and multiple well-regarded PEFT baselines. The descriptions of our selective baselines are as follows: 
\begin{itemize}
    \item \textbf{Full fine-tuning (FFT)} initializes the base model with pre-trained weights and biases, updating all parameters during fine-tuning.
    Full fine-tuning typically serves as a comparative performance upper bound for PEFT methods \citep{valipour2023dylora}.
    \item \textbf{Bitfit}~\citep{zaken2021bitfit} fine-tunes the bias vectors, leaving other model parameters unchanged.
    \item \textbf{Adapter tuning} integrates tunable adapter layers into Transformer blocks, featuring a pair of down-projection and up-projection matrices with a non-linear activation function in between. We compare four Adapter variants:
    $\textbf{Adapter}^{\textbf{H}}$ \citep{houlsby2019parameter} inserts adapter layers after the attention and the feed-forward block to fine-tune.
    $\textbf{Adapter}^{\textbf{D}}$ \citep{ruckle2021adapterdrop} discards non-activated adapters to improve fine-tuning efficiency.
    $\textbf{Adapter}^{\textbf{L}}$ \citep{lin2020exploring} employs an efficient design, placing adapter layers after the MLP module and LayerNorm.
    $\textbf{Adapter}^{\textbf{P}}$ \citep{pfeiffer2021adapterfusion} applies adapter after the feed-forward layer and employs a two-stage learning strategy to enhance multi-task performance. 
    \item \textbf{LoRA} \citep{hu2021lora} only fine-tunes a pair of low-rank matrices to approximate the task-specific knowledge updates, effectively diminishing the number of trainable parameters.
    \item \textbf{AdaLoRA} \citep{zhang2022adaptive} reparameterizes task-specific knowledge updates in the SVD form and adaptively allocates the parameter budget through pruning the less important singular values.
    \item \textbf{DyLoRA} \citep{valipour2023dylora} dynamically trains LoRA for a range of ranks, reducing the training time to find a fixed, optimal rank.
    \item \textbf{VeRA} \citep{kopiczko2023vera} employs learnable vectors to adapt a shared pair of frozen random matrices across layers to reduce the trainable parameters count.
    \item \textbf{DoRA} \citep{liu2024dora} decomposes
    the base model weights into magnitude and direction components for fine-tuning, reducing the number of trainable parameters.
    \item \textbf{PiSSA} \citep{meng2024pissa} performs SVD to portion the base model into principal components with larger singular values and residual components with smaller ones, fine-tuning the low-rank matrices initialized with the principle components while keeping the residual components unchanged.
    \item \textbf{MiLoRA} \citep{wang2024milora} also utilizes SVD for parameter initialization but diverges from PiSSA by fine-tuning low-rank matrices initialized with residual components and maintaining the principal ones unchanged.
    \item \textbf{SARA} \citep{gu2024sara} conducts SVD at the initialization stage to adaptively find the appropriate rank for each layer.
    \item \textbf{CorDA} \citep{yangcorda} performs SVD on the base model, oriented by the covariance matrix that encodes the context of the target task. CorDA supports two fine-tuning modes: 1) initializing the tunable low-rank matrices with principal components for enhanced performance; and 2) freezing the principle components while using minor components to initialize tunable matrices, thereby preserving world knowledge.
\end{itemize}

\section{Details of Benchmark Datasets}
\subsection{GLUE Benchmark}\label{sec:glue_benchmark}
For natural language understanding (NLU), we employ the GLUE benchmark \citep{wang2018glue}, which is a widely used benchmark containing a collection of 8 NLU datasets, including CoLA, SST-2, MRPC, STS-B, QQP, MNLI, QNLI, and RTE. We present the statistical information of the GLUE benchmark in the table below.

\begin{table*}[ht!]
\centering
\caption{Overview of task descriptions and dataset statistics within the GLUE benchmark.
}
\label{tab:data-glue}
\resizebox{\linewidth}{!}{%
\begin{tabular}{llrrrrll}
\toprule
\multicolumn{1}{l|}{\textbf{Corpus}} & \textbf{Task} & \textbf{\# Train} & \textbf{\# Val} & \textbf{\# Test} & \textbf{\# Labels} & \textbf{Metrics} & \textbf{Domain} \\ 
\midrule
\multicolumn{8}{c}{Single-Sentence Tasks} \\ 
\midrule
\multicolumn{1}{l|}{CoLA} & Acceptability & 8.55k & 1.04k & 1.06k & 2 & Matthews Corr. & misc. \\
\multicolumn{1}{l|}{SST-2} & Sentiment & 67.3k & 872 & 1.82k & 2 & Accuracy & Movie reviews \\ 
\midrule
\multicolumn{8}{c}{Similarity and Paraphrase Tasks} \\ 
\midrule
\multicolumn{1}{l|}{MRPC} & Paraphrase & 3.67k & 408 & 1.73k & 2 & Accuracy/F1 & News \\
\multicolumn{1}{l|}{STS-B} & Sentence similarity & 5.75k & 1.5k & 1.38k & 1 & Pearson/Spearman Corr. & misc. \\
\multicolumn{1}{l|}{QQP} & Paraphrase & 364k & 40.4k & 391k & 2 & Accuracy/F1 & Social QA \\ \midrule
\multicolumn{8}{c}{Inference Tasks} \\ 
\midrule
\multicolumn{1}{l|}{MNLI} & NLI & 393k & 19.65k & 19.65k & 3 & Accuracy & misc. \\
\multicolumn{1}{l|}{QNLI} & QA/NLI & 105k & 5.46k & 5.46k & 2 & Accuracy & Wikipedia \\
\multicolumn{1}{l|}{RTE} & NLI & 2.49k & 277 & 3k & 2 & Accuracy & News \& Wikipedia \\ 
\bottomrule
\end{tabular}%
}
\end{table*}

\subsection{E2E NLG Challenge}\label{sec:e2e_benchmark}
For natural language generation (NLG), we utilize the E2E (End-to-End) NLG Challenge dataset \citep{novikova2017e2e}, which is commonly used for the evaluation of natural language generation models. 
This dataset includes approximately 42k training samples, 4.6k validation samples, and 4.6k test samples from the restaurant domain. 
The E2E dataset involves evaluations across five metrics: BLEU, NIST, METEOR, ROUGE-L, and CIDEr. Detailed explanations of these metrics are as follows:

\begin{itemize}
    \item \textbf{BLEU} (Bilingual Evaluation Understudy) evaluates the quality of machine-generated text by comparing it to one or more human-generated reference translations.
    \item \textbf{NIST} (National Institute of Standards and Technology) evaluates the quality of machine-generated text by calculating the similarity between a machine output and a reference text using weighted average of n-grams precision.
    \item \textbf{METEOR} (Metric for Evaluation of Translation with Explicit ORdering) measures the alignment between the machine-generated and reference texts by calculating a score based on the harmonic mean of precision and recall.
    \item \textbf{ROUGE-L} (Recall-Oriented Understudy for Gisting Evaluation) measures the longest common subsequence(LCS) between the machine output and the reference.
    It specifically focuses on the sequence of words, making it sensitive to the fluency and order of information in the generated text.
    \item \textbf{CIDEr} (Consensus-based Image Description) measures the similarity of the machine-generated text and the human-generated ground truth by considering both the n-gram overlap and the consensus among human annotators. 
\end{itemize}

\subsection{Synthetic Dataset}\label{sec:synthetic_dataset}
For instruction following tasks, we employ synthetic datasets generated using \revise{GPT4o}, based on the foundational ``No Robots'' seed dataset \citep{no_robots}.
Task-specific subsets, including summarization, classification, coding, and closed QA, serve as seeds for generating synthetic data through the framework proposed by \citep{park2024llamaduo}.
Table \ref{tab:seed_dataset_volume} presents the volume of data samples and token-level statistical information for these task-specific synthetic subsets.

\begin{table}[ht!]
\centering
\caption{Data volume and token-level statistics of the train and test synthetic datasets generated by GPT4o for each instruction-following task.}
\label{tab:seed_dataset_volume}
\resizebox{0.7\linewidth}{!}{
\begin{tabular}{l|l|cccccc} 
\toprule
\multirow{2}*{\textbf{Task}} & \multirow{2}*{\textbf{Split}} & \multicolumn{2}{c}{\textbf{Data Volume}} & \multicolumn{4}{c}{\textbf{Token-level Statistics}} \\ 
\cmidrule(lr){3-4} \cmidrule(lr){5-8}
& & \textbf{Seed} & \textbf{Synthesis} & \textbf{Min} & \textbf{Max} & \textbf{Avg.} & \textbf{Std.} \\ 
\midrule
\multirow{2}{*}{Summarization} & Train & 395 & 128K & 10 & 2,386 & 95 & 53 \\
& Test  & 25  & 100  & 148 & 1,150 & 426 & 245 \\
\midrule
\multirow{2}*{Classification} & Train & 334 & 128K & 6 & 2,159 & 67 & 37 \\
& Test  & 16  & 64  & 46 & 520 & 119 & 109 \\
\midrule
\multirow{2}*{Coding} & Train & 334 & 128K & 9 & 6,518 & 151 & 84 \\
& Test  & 16  & 64  & 49 & 821 & 317 & 189 \\
\midrule
\multirow{2}*{Closed QA} & Train & 245 & 128K & 12 & 1,701 & 135 & 59\\
& Test  & 15  & 60  & 126 & 1,578 & 411 & 378 \\
\bottomrule
\end{tabular}
}
\vspace{-10pt}
\end{table}

\subsection{Alpaca and MT-Bench} \label{sec:alpaca_MTBench}
\textbf{Alpaca} \citep{alpaca} is a well-known instruction dataset that contains 51k instruction-following demonstrations generated by text-davinci-003. 
These data are synthesized using an improved self-instruct method \citep{wang2023self}. The dataset is designed for instruction-tuning LLMs to improve their ability to follow instructions. Each sample includes an instruction, an input (if applicable), and an output. A specific example is presented below.

\begin{minipage}{\textwidth}
\begin{lstlisting}[language=python]
{
  "instruction": "Create a classification task by clustering the given list of items.",
  "input": "Apples, oranges, bananas, strawberries, pineapples",
  "output": "Class 1: Apples, Oranges\nClass 2: Bananas, Strawberries\nClass 3: Pineapples",
  "text": "Below is an instruction that describes a task, paired with an input that provides further context. Write a response that appropriately completes the request.\n\n### Instruction:\nCreate a classification task by clustering the given list of items.\n\n### Input:\nApples, oranges, bananas, strawberries, pineapples\n\n### Response:\nClass 1: Apples, Oranges\nClass 2: Bananas, Strawberries\nClass 3: Pineapples"
}
\end{lstlisting}
\end{minipage}

The \texttt{\textbf{instruction}} describes the targeted task to be performed by the model. Each of the 52k instructions is unique. 
The \texttt{\textbf{input}} can represent the optional input to the task or serve as the additional context to the corresponding instruction. 
The \texttt{\textbf{output}} is the response generated by text-davinci-003 to the associated instruction.
The \texttt{\textbf{Text}} is the formatted combination of the instruction, input, and output, using the prompt template for fine-tuning models.

\textbf{MT-bench} \citep{zheng2023judging} contains 80 predefined open-ended questions across diverse domains such as writing, roleplay, reasoning, math, coding, extraction, STEM, and humanities.
These challenging questions are designed to automatically assess an LLM's instruction-following capabilities, with advanced service LLMs like GPT-4 acting as judges.
Below is an example from MT-bench.

\begin{minipage}{\textwidth}
\begin{lstlisting}[language=python]
{
  "question_id": 101,
  "category": "reasoning",
  "turns": [
    "Imagine you are participating in a race with a group of people. If you have just overtaken the second person, what's your current position? Where is the person you just overtook?",
    "If the \"second person\" is changed to \"last person\" in the above question, what would the answer be?"
  ],
  "reference": [
    "You are in second place.",
    "Uncertain."
  ]
}
\end{lstlisting}
\end{minipage}

\subsection{Commonsense Reasoning}
The Commonsense170K dataset \citep{hu2023llm} contains data samples from eight well-known commonsense reasoning tasks: 
\begin{itemize}
    \item \textbf{BoolQ} \citep{clark2019boolq} dataset comprises 15,942 naturally occurring yes/no questions, generated in unprompted and unconstrained settings. 
    \item \textbf{PIQA} \citep{bisk2020piqa} dataset consists of samples structured as multiple-choice questions, each presenting a question with two possible solutions that require physical commonsense to answer.
    \item \textbf{SIQA} \citep{sap2019socialiqa} dataset contains multiple-choice questions regarding the pragmatic implications of social events, which can measure LLMs' abilities to address social commonsense reasoning.
    \item \textbf{HellaSwag} \citep{zellers2019hellaswag} dataset includes commonsense natural language inference questions, offering a context and multiple endings to complete it.
    \item \textbf{WinoGrande} \citep{sakaguchi2021winogrande} dataset is structured as a fill-in-the-blank task with two options, designed to test a model's ability to correctly solve the problem using commonsense reasoning.
    \item \textbf{ARC-e} and \textbf{ARC-c} are the Easy and Challenge Set of the ARC \citep{clark2018think} dataset, which contains grade-school level, multiple-choice science questions. Notably, the Challenge Set includes questions answered incorrectly by both the retrieval-based algorithm and word co-occurrence algorithm.
    \item \textbf{OBQA} \citep{mihaylov2018can} dataset contains multiple-choice elementary-level science questions requiring multi-step reasoning, use of additional common and provided science facts (open book), and rich text comprehension.
\end{itemize}

\section{Prompt Templates}\label{sec:prompt_templates}
\begin{figure}[h]
\centering
\includegraphics[width=\textwidth]{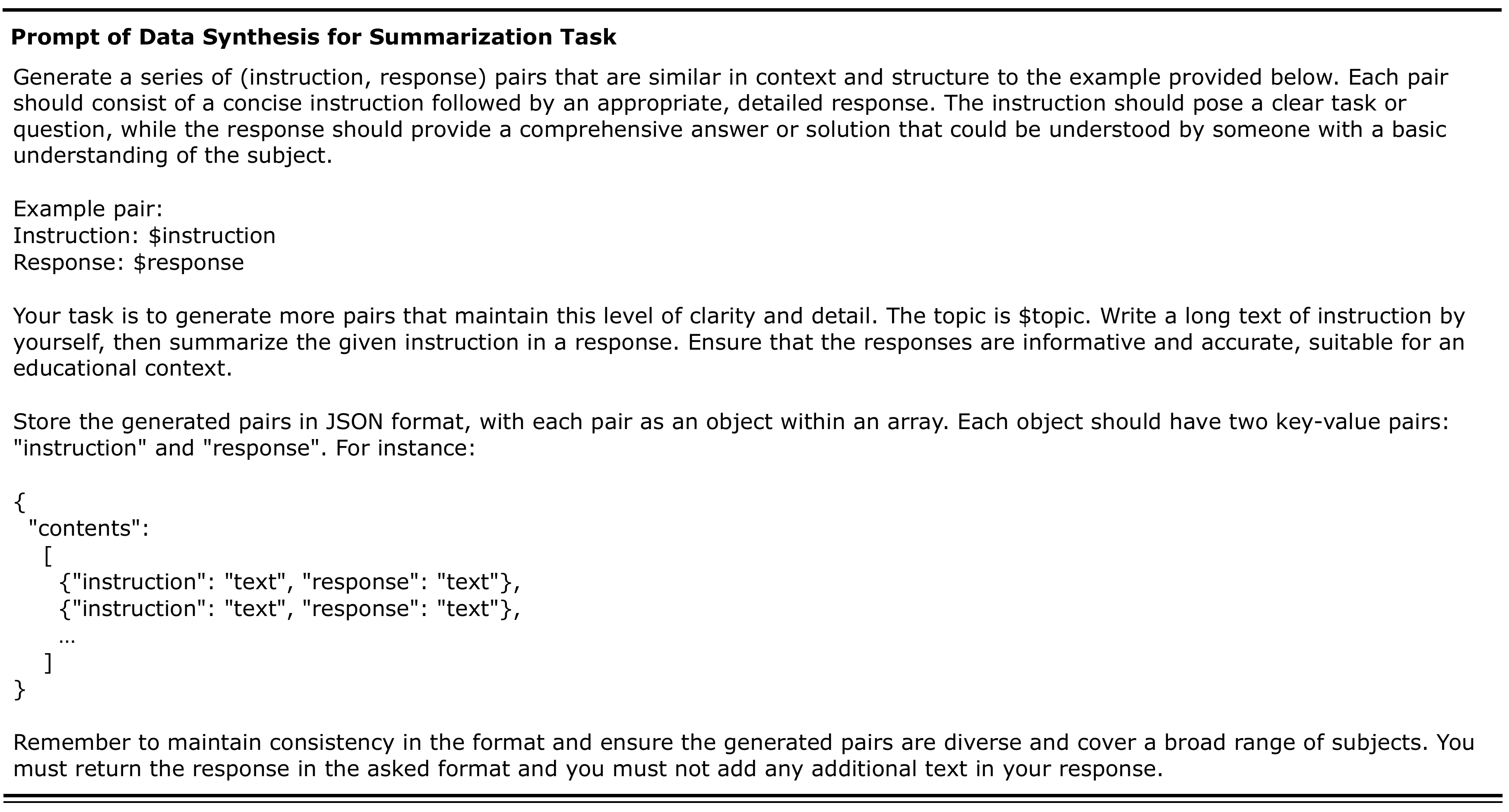}
\caption{Prompt template of data synthesis for summarization tasks by GPT4o.}
\label{fig:default-prompt-synthegen_one_task}
\end{figure}

\begin{figure}[h]
\centering
\includegraphics[width=\textwidth]{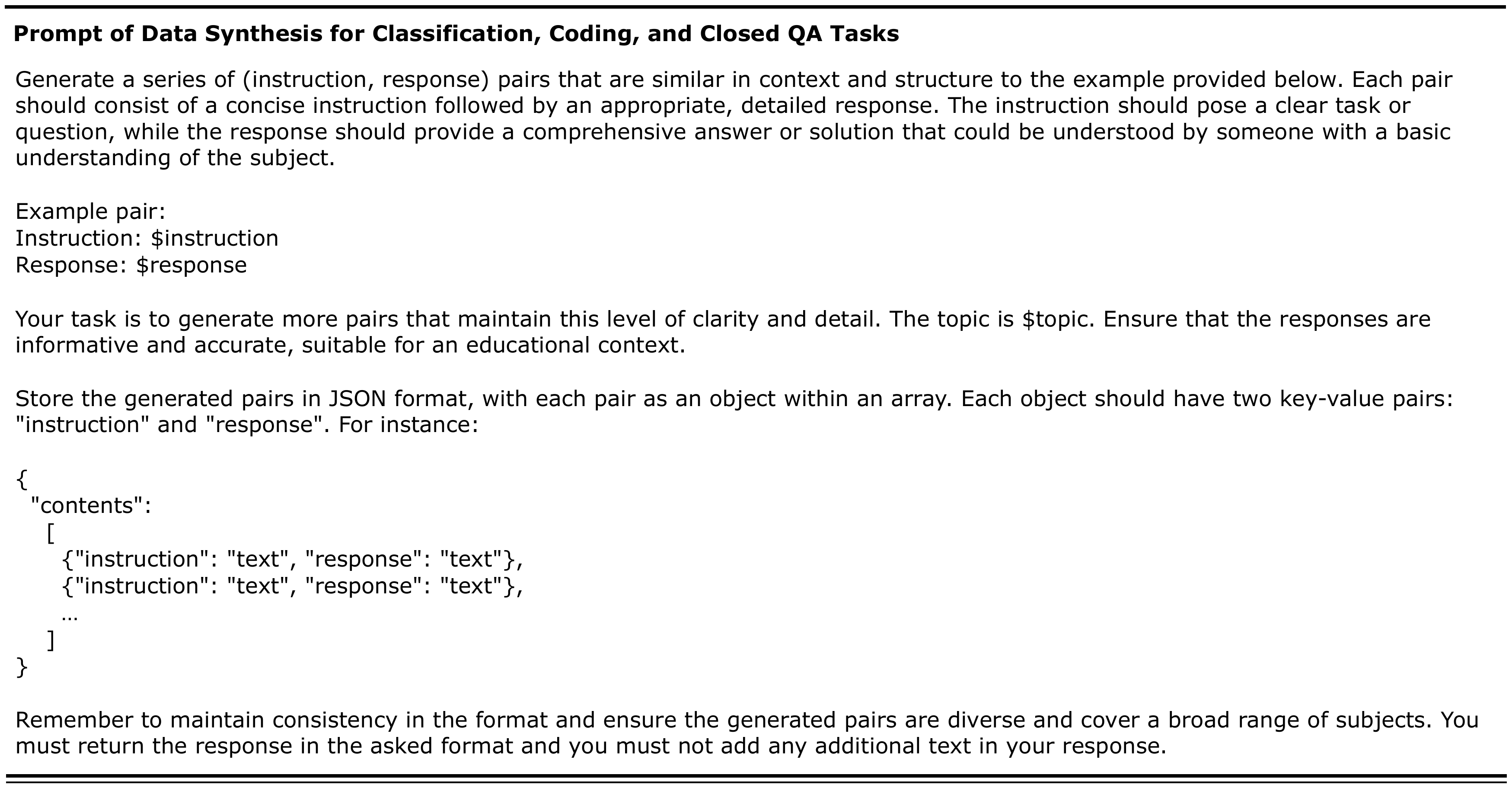}
\caption{Prompt template of data synthesis for classification, coding, and closed QA tasks by GPT4o.}
\label{fig:default-prompt-synthegen_four_tasks}
\end{figure}

\begin{figure}[h]
\centering
\includegraphics[width=\textwidth]{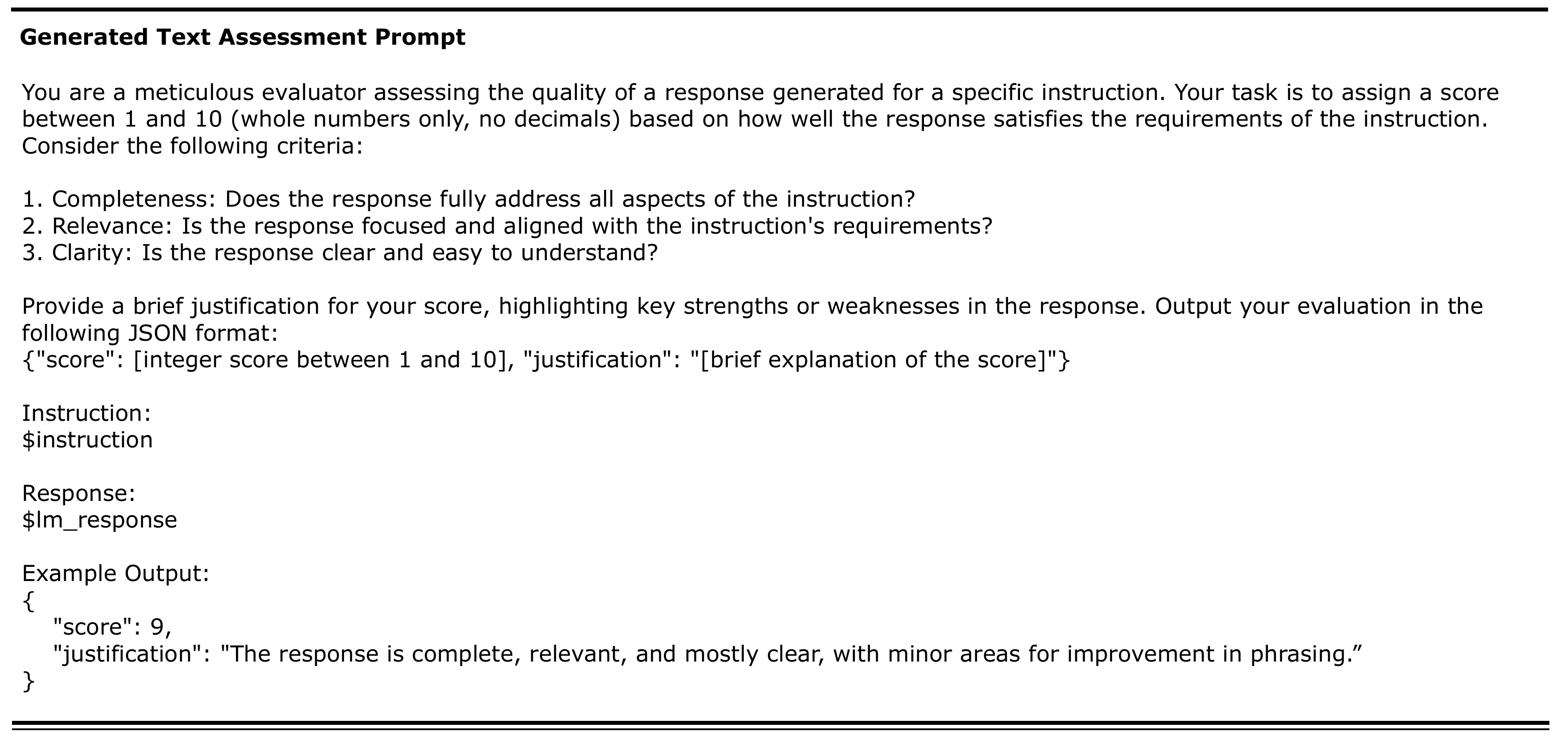}
\caption{Prompt template to evaluate the fine-tuned model's response by GPT4o.}
\label{fig:default-prompt-assessment}
\end{figure}

Following the typical practices of \citep{wang2023self} and \citep{zheng2023judging}, we leverage two specialized prompt templates: 1) one for generating synthetic datasets and 2) another for evaluating the outputs of fine-tuned LLMs. 
To be specific, Figure \ref{fig:default-prompt-synthegen_one_task} presents the prompt template crafted for generating synthetic data aimed at the summarization task, whereas Figure \ref{fig:default-prompt-synthegen_four_tasks} shows the prompt template for other tasks. 
We guide GPT4o in generating analogous data samples by using a reference example pair consisting of a prompt \texttt{\$instruction} and its corresponding response \texttt{\$response} from the training subset of the seed dataset.
In addition, the template is designed to request multiple synthetic data samples in a single query, thus maximizing the efficiency of API use.
On the other hand, Figure \ref{fig:default-prompt-assessment} shows the prompt template used for assessing the precision and similarity between the response \texttt{\$lm\_response} and \texttt{\$human\_response}
given the same \texttt{\$instruction} from the test subset of the seed dataset, where the \texttt{\$} symbol indicates a placeholder, designed to be substituted with actual data during the runtime. 
We only report the precision results in our experiments for the sake of brevity.
Given the unique features of different downstream tasks, there is no optimal prompt template that universally applies. 
Therefore, the actual content of the prompt template is adjusted to align with the specific requirements of the task for which the synthetic dataset is being generated.

\section{Training Details}\label{sec:train_details}
\subsection{Natural Language Understanding}\label{sec:details_nlu}
For NLU tasks, we align with the experimental setup detailed in \citep{hu2021lora,zhang2022adaptive} for a fair comparison. 
The detailed configurations of KaSA for RoBERTa-base, RoBERTa-large, and DeBERTaV3-base on the GLUE benchmark are depicted in Table \ref{tab:hyper_roberta} and Table \ref{tab:hyper_deberta}, respectively.
It is important to note that our adaptation process for the MRPC, RTE, and STS-B tasks begins with the pre-trained RoBERTa model, rather than a model that has already been adapted to MNLI. 
As a result, we fine-tune the models on all datasets starting from their original pre-trained weights. 
The results we present are the median results from 5 runs, each conducted with a distinct random seed.

\begin{table}[t]
    \centering
    \caption{The hyperparameters we used for RoBERTa-base and RoBERTa-large on the GLUE benchmark.}
    \resizebox{\linewidth}{!}{%
    \begin{tabular}{ll|cccccccc}
        \toprule
        Model  & Settings     & MNLI & SST-2 & MRPC & CoLA & QNLI & QQP & RTE & STS-B \\
        \midrule
        \multirow{3}{*}{Common}     & Optimizer   & \multicolumn{8}{c}{AdamW} \\
                              & Warmup Ratio & \multicolumn{8}{c}{0.06} \\
                              & LR Schedule & \multicolumn{8}{c}{Linear} \\
        \midrule
        \multirow{10}{*}{\makecell{RoBERTa$_\text{base}$}} \ 
                              & Batch Size & 32 & 128 & 32 & 32 & 32 & 128 & 32 & 32 \\
                              & \# Epochs & 100 & 100 & 100 & 100 & 10 & 100 & 100 & 40 \\
                              & Learning Rate & 5E-04 & 5E-04 & 4E-04 & 4E-04 & 4E-04 & 5E-04 & 4E-04 & 3E-04 \\
                              & Weight Decay & 0.0 & 0.0 & 0.0 & 0.0 & 0.0 & 0.0 & 0.0 & 0.0   \\
                              & KaSA Rank & \multicolumn{8}{c}{$r_{query}=r_{value}=8$} \\
                              & KaSA $\alpha$ & \multicolumn{8}{c}{16} \\
                              & KaSA $\beta$ & 2.4E-3 & 1E-04 & 1E-01 & 1E-04 & 1E-02 & 1E-4 & 2.4E-01 & 1E-04 \\
                              & KaSA $\gamma$  & 2.4E-4 & 1E-03 & 1E-03 & 1E-03 & 1E-05 & 1E-3 & 2.4E-04 & 1E-05  \\
                              & KaSA Dropout   & 0.0 & 0.0 & 0.0 & 0.0 & 0.0 & 0.0 & 0.0 & 0.0   \\
                              & Max Seq. Len. & 512 & 512 & 512 & 512 & 512 & 512 & 512 & 512 \\
        \midrule
        \multirow{10}{*}{\makecell{RoBERTa$_\text{large}$}} \ 
                              & Batch Size & - & 64 & 32 & 32 & 8 & - & 32 & 32 \\
                              & \# Epochs & - & 10 & 10 & 100 & 20 & - & 100 & 20 \\
                              & Learning Rate & - & 4E-04 & 3E-04 & 3E-04 & 4E-04 & - & 4E-04 & 3E-04 \\
                              & Weight Decay & - & 0.1 & 0.1 & 0.0 & 0.0 & - & 0.0 & 0.0   \\
                              & KaSA Rank & \multicolumn{8}{c}{$r_{query}=r_{value}=8$} \\
                              & KaSA $\alpha$ & \multicolumn{8}{c}{16} \\
                              & KaSA $\beta$ & - & 1E-04 & 1E-02 & 2.4E-01 & 1E-02 &- & 1E-04 & 1E-03 \\
                              & KaSA $\gamma$  &- & 1E-04 & 1E-02 & 2.4E-04 & 1E-03 &- & 1E-03 & 1E-02  \\
                              & KaSA Dropout   &- & 0.0 & 0.0 & 0.0 & 0.0 & - & 0.0 & 0.0   \\
                              & Max Seq. Len. &  -& 512 & 512 & 512 & 512 & - & 512 & 128 \\
        \bottomrule
    \end{tabular}}
    \label{tab:hyper_roberta}
\end{table}

\begin{table}[t] 
\centering 
\caption{The hyperparameters we used for DeBERTaV3-base on the GLUE benchmark.}
\resizebox{0.9\textwidth}{!}{%
\begin{tabular}{ll|cccccc}
\toprule
 Model & Settings      & SST-2    & MRPC & CoLA     & QNLI     & RTE    & STS-B       \\ 
\midrule
        & Optimizer  & \multicolumn{6}{c}{AdamW}   \\ 
        & Warmup Ratio  & \multicolumn{6}{c}{0.06}   \\ 
        & LR Scheduler   & \multicolumn{6}{c}{Linear}         \\  
\midrule
\multirow{10}*{DeBERTaV3-base} & Batch size   & 128 & 32 & 32 & 16 & 32 &  32        \\ 
                        & \# Epochs       & 10   & 10   &100   &20  & 100  & 20       \\ 
                        & Learning Rate   & 5E-4 & 4E-4 &4E-4 & 4E-4   & 5E-4  & 4E-4  \\ 
                        & Weight Decay    & 0.0 & 0.0 & 0.0 & 0.0 & 0.0 & 0.0  \\ 
                        & KaSA Rank   & \multicolumn{6}{c}{$r_{query} = r_{value} = 8$}         \\
                        & KaSA $\alpha$   & \multicolumn{6}{c}{16}         \\
                        & KaSA $\beta$  & 1E-04 & 1.0 & 2.4E-01 & 1E-01 & 1E-04 & 1E-01  \\
                        & KaSA $\gamma$  & 1E-03 & 1.0 & 2.4E-04 & 1E-01 & 1E-03 & 1E-01 \\
                        & KaSA Dropout   & 0.0 & 0.0 & 0.0 & 0.0 & 0.0 & 0.0   \\
                        & Max Seq. Len.  & 512   &512  &64  &512 & 512 & 512    \\  
\bottomrule
\end{tabular}
}
\label{tab:hyper_deberta}
\end{table}

\subsection{Natural Language Generation}\label{sec:training_nlg}
For NLG tasks, our KaSA adheres to the experimental setup outlined in \citep{hu2021lora,gu2024sara} to ensure a fair comparison. 
The comprehensive configurations of KaSA for GPT-2 Medium and GPT-2 Large models on the E2E NLG Challenge benchmark are depicted in Table \ref{tab:hyper_gpt2}.

\begin{table}[h]
\centering
\caption{The hyperparameters for GPT-2 on E2E NLG Challenge.}
\resizebox{0.5\textwidth}{!}{%
\begin{tabular}{ll|cc}
\toprule
Stage & Settings & Medium & Large  \\
\midrule
\multirow{13}*{Training} & Optimizer & \multicolumn{2}{c}{AdamW} \\
& Weight Decay & 0.01 & 0.01 \\
& Dropout Prob & 0.1 & 0.1 \\
& Batch Size & \multicolumn{2}{c}{8} \\
& \# Epoch & \multicolumn{2}{c}{5} \\
& Warmup Steps & \multicolumn{2}{c}{500} \\
& LR Scheduler & \multicolumn{2}{c}{Linear} \\
& Label Smooth & 0.1 & 0.1 \\
& Learning Rate & \multicolumn{2}{c}{2E-4} \\
& KaSA Rank & \multicolumn{2}{c}{$r_{query}=r_{value}=4$} \\
& KaSA $\alpha$ & \multicolumn{2}{c}{32} \\
& KaSA $\beta$ & \multicolumn{2}{c}{1E-4}  \\
& KaSA $\gamma$  & \multicolumn{2}{c}{1E-3}   \\
\midrule
\multirow{3}*{Inference} & 
Beam Size & \multicolumn{2}{c}{10} \\
& Length Penalty & 0.9 & 0.8 \\
& no repeat ngram size & \multicolumn{2}{c}{4} \\
\bottomrule
\end{tabular}
}
\label{tab:hyper_gpt2}
\end{table}

\subsection{Instruction Following}\label{sec:instruction_following_config}
For instruction following tasks, we adopt the framework proposed by \citep{park2024llamaduo} to streamline the processes of data synthesis, fine-tuning, and evaluation. 
We fine-tune several of the most popular LLMs, including LLaMA3 8B, Mistal 7B, Gemma 7B, and LLaMA2 13B, using KaSA and different PEFT baselines to facilitate comparative analysis. 
Detailed hyperparameter configurations are provided in Table \ref{tab:training-configurations}.    

\begin{table}[t] 
    \centering 
    \caption{Detailed configurations used for the instruction following task.}
    \resizebox{0.98\textwidth}{!}{%
    \begin{tabular}{ll|ccccc}
    \toprule
    Stage & Settings   & Classification   & Summarization  & Coding  & Closed QA & MT-Bench  \\ 
    \midrule
    \multirow{13}*{Training} & Optimizer & \multicolumn{5}{c}{AdamW} \\
             & Batch Size & \multicolumn{5}{c}{Gemma 7B $=$ 8, Mitral 7B $=$ LLaMA3 8B $=$ 16} \\
             & \# Epoch & \multicolumn{5}{c}{1} \\
             & Warmup Ratio & \multicolumn{5}{c}{0.1} \\
              & Data Type  & \multicolumn{5}{c}{Bfloat16}   \\ 
             & LR Scheduler   & \multicolumn{5}{c}{Cosine}       \\  
             & Learning Rate   & \multicolumn{5}{c}{2.0E-04}     \\ 
             & KaSA Rank   & \multicolumn{5}{c}{$r_{query} = r_{value} = 8$}         \\
             & KaSA $\alpha$   & \multicolumn{5}{c}{16}         \\
             & KaSA $\beta$ & \multicolumn{5}{c}{1E-4}   \\
             & KaSA $\gamma$  & \multicolumn{5}{c}{1E-3}   \\
             & KaSA Dropout    & \multicolumn{5}{c}{0.05}     \\ 
             & Max Seq. Len.  & \multicolumn{5}{c}{512}     \\  
    \midrule
    \multirow{3}*{Inference} & 
        Number of Beams & \multicolumn{5}{c}{10}  \\
        & Length Penalty & \multicolumn{5}{c}{0.8}  \\
        & No Repeat N-Gram Size & \multicolumn{5}{c}{4}  \\
             \bottomrule
    \end{tabular}}
    \label{tab:training-configurations}
\end{table}

\subsection{Commonsense Reasoning}\label{sec:commonsense_reasoning}
We adhere strictly to the hyperparameter configurations for training and evaluation as specified by \citep{wang2024milora} and \citep{hu2023llm}, \textbf{without any tuning}. 
The specific hyperparameter configurations used are shown in Table \ref{tab:commonsense_reasoning_hyper}.

\begin{table}[h]
\centering
\caption{The hyperparameter configurations for LLaMA2 7B and LLaMA3 8B on commonsense reasoning tasks. To ensure a fair comparison, these configurations remain consistent across LoRA, PiSSA, and MiLoRA, with the exception of the specific hyperparameters unique to KaSA, namely $\beta$ and $\gamma$, as well as PiSSA and MiLoRA, where $\alpha=32$.
}
\resizebox{0.55\textwidth}{!}{%
\begin{tabular}{lcc}
\toprule
\multirow{2}*{Hyperparameters} & \multicolumn{2}{c}{Commonsense Reasoning}  \\
\cmidrule(lr){2-3}
& LLaMA2 7B & LLaMA3 8B \\
\midrule
Optimizer & \multicolumn{2}{c}{AdamW} \\
Batch Size & \multicolumn{2}{c}{16} \\
\# Epoch & \multicolumn{2}{c}{3} \\
Warmup Steps & \multicolumn{2}{c}{100} \\
LR Scheduler & \multicolumn{2}{c}{Linear} \\
Learning Rate & \multicolumn{2}{c}{3E-4} \\
KaSA Rank & \multicolumn{2}{c}{32} \\
KaSA $\alpha$ & \multicolumn{2}{c}{64} \\
Dropout Prob & \multicolumn{2}{c}{0.05} \\
KaSA $\beta$ & 1E-2 & 1E-4  \\
KaSA $\gamma$  & 1E-3 & 1E-3 \\
Placement & \multicolumn{2}{c}{query, key, value, MLP up, MLP down} \\
\bottomrule
\end{tabular}
}
\label{tab:commonsense_reasoning_hyper}
\end{table}

\section{Additional Experimental Results}
\subsection{Natural Language Understanding on DeBERTaV3-base}
\label{sec:addition_nlu}
\begin{table}[t]
\centering
\caption{Performance of DeBERTaV3-base ($\text{DeB}_\text{v3}$) with different adaptation methods on 6 datasets of the GLUE benchmark.  
We report the average result of five runs with different random seeds.
The best results for each dataset are shown in \textbf{bold}.
Higher is better for all metrics.
}
\label{tab:glue_datasets_deberta}
\resizebox{0.85\linewidth}{!}{
\begin{tabular}{l|r|cccccccc}
\toprule
\multirow{2}*{\textbf{ Model(Method)}} & \textbf{\multirow{2}*{\makecell[r]{\# Trainable \\Parameters}}} & {\textbf{ SST-2}} & {\textbf{ MRPC}}  & {\textbf{ CoLA}} & {\textbf{ QNLI}} & {\textbf{ RTE}} & {\textbf{ STS-B}} & {\textbf{ All}} \\
~ & ~ & {(Acc.)} & {(Acc.)} &  {(Mcc.)} & {(Acc.)} & {(Acc.)} & {(Pcc.)} & {Avg.} \\
\midrule 
$\text{DeB}_\text{v3}$(FFT) & 184.0M        & 95.63     & 89.46        & 69.19     & 94.03     & 83.75   & 91.60   & 87.28    \\ 
$\text{DeB}_\text{v3}$($\text{Adpt}^{\text{H}}$) &0.6M        & 95.30     & 89.22        & 67.87     & 93.76     & 85.56   & 91.30   & 87.17    \\ 
$\text{DeB}_\text{v3}$($\text{Adpt}^{\text{P}}$) &0.6M     & 95.53     & 89.22        & 69.48     & 93.98      & 84.12   & 91.52   & 87.31    \\ 
$\text{DeB}_\text{v3}$(LoRA) &0.3M      & 94.95     & 89.71        & 68.71     & 94.03    & 85.56   & \textbf{91.68}   & 87.44     \\ 
$\text{DeB}_\text{v3}$(AdaLoRA) &0.3M        & 95.80     & 90.44         & 70.04    & {94.49}  & 87.36   & 91.63   & 88.29    \\ 
$\text{DeB}_\text{v3}$($\text{PiSSA}$)  & 0.3M  & 95.30 & \textbf{91.42} & 70.29 & 93.59 & 84.84 &  91.37 & 87.80 \\
$\text{DeB}_\text{v3}$($\text{MiLoRA}$)  & 0.3M  & 95.99 &  89.71 & 70.34 & {94.14} & 85.92 & 90.28  & 87.73 \\
$\text{DeB}_\text{v3}$($\text{KaSA}$)  & 0.3M & \textcolor{black}{\textbf{96.22}} & \textcolor{black}{\textbf{91.42}} & \textcolor{black}{\textbf{70.41}} & \textcolor{black}{\textbf{94.55}}  & \textcolor{black}{\textbf{88.09}} & \textcolor{black}{91.62} & \textbf{88.72}\\
\bottomrule
\end{tabular}
}
\end{table}
As demonstrated in Table \ref{tab:glue_datasets_deberta}, the DeBERTaV3-base results consistently surpass all baseline performances across the datasets, with the exception of STS-B, achieving the highest average performance of 88.72\%. This further validates the efficacy of our method across different model architectures.

\subsection{Natural Language Generation}
\label{sec:addition_nlg}
\textbf{Models and Datasets.}
For NLG tasks, we employ KaSA and other PEFT baselines to fine-tune both GPT-2 Medium (355M) and GPT-2 Large (774M) models \citep{radford2019language} on the well-established E2E (End-to-End) NLG Challenge benchmark \citep{novikova2017e2e}, \revise{which focuses on restaurant domain information.}
\revise{The statistics of the E2E NLG Challenge benchmark and the evaluation metrics applied are detailed in \ref{sec:e2e_benchmark}.}

\textbf{Implementation Details.}
We adopt the experimental configurations delineated in \citep{hu2021lora,gu2024sara} for the fine-tuning of \texttt{query} and \texttt{value} weights within each Transformer block, setting a rank of $r = 4$. The AdamW optimizer is employed, paired with a linear learning rate schedule over 5 epochs. 
The reported results represent the mean outcomes from 3 runs, each initialized with a distinct random seed, selecting the performance at the last epoch of each run for comparison.
For further details on the hyperparameters utilized, refer to \ref{sec:training_nlg}.

\textbf{Main Results.}
\begin{table}[t]
\centering
\caption{Performance of GPT-2 Medium and Large models with different adaptation methods on the E2E NLG Challenge.
For all metrics, higher values indicate better performance. $^*$ indicates that the results are reported in prior works. Best results are shown in \textbf{bold}.
}
\label{tab:gpt2_e2e}
\resizebox{0.85\linewidth}{!}{
\begin{tabular}{l|r|ccccc}
\toprule
\textbf{Model(Method)} & \textbf{\makecell[r]{\# Trainable \\Parameters}}  & \textbf{BLEU} & \textbf{NIST} & \textbf{METEOR} & \textbf{ROUGE-L} & \textbf{CIDEr} \\
\midrule
$\text{GPT-2}_\text{Medium}$(FFT*)                      & 354.92M  & 68.2   & 8.62   & 46.2 & 71.0 
& 2.47         \\ 
$\text{GPT-2}_\text{Medium}$($\text{Adpt}^{\text{L}}$*) &   0.37M  & 66.3   & 8.41   & 45.0 & 69.8     & 2.40         \\ 
$\text{GPT-2}_\text{Medium}$($\text{Adpt}^{\text{L}}$*) &  11.09M  & 68.9   & 8.71   & 46.1 & 71.3     & 2.47         \\ 
$\text{GPT-2}_\text{Medium}$($\text{Adpt}^{\text{H}}$*) &  11.09M  & 67.3  & 8.50   & 46.0 & 70.7    & 2.44        \\  
$\text{GPT-2}_\text{Medium}$(LoRA*) &   0.35M  & 70.4   & 8.85 & 46.8 & 71.8    & 2.53         \\ 
$\text{GPT-2}_\text{Medium}$(AdaLoRA) & 0.38M & 68.2 & 8.58 & 44.1 & 70.7 & 2.35\\ 
$\text{GPT-2}_\text{Medium}$(DyLoRA) & 0.39M & 69.2 & 8.75 & 46.3 & 70.8 & 2.46\\ 
$\text{GPT-2}_\text{Medium}$(VeRA) & 0.098M & 69.1 & 8.71 & 46.3 & 70.8 & 2.43 \\ 
$\text{GPT-2}_\text{Medium}$(SARA) & 0.33M & 70.4 & 8.84 & 46.7 & \textbf{72.3} & \textbf{2.55} \\  
$\text{GPT-2}_\text{Medium}$(KaSA) & 0.35M & \textbf{70.6} & \textbf{8.86} & \textbf{46.9}  & 72.1 & \textbf{2.55} \\ 
\midrule 
$\text{GPT-2}_\text{Large}$(FFT*)  & 774.03M  & 68.5   & 8.78   & 46.0 & 69.9     & 2.45         \\ 
$\text{GPT-2}_\text{Large}$($\text{Adpt}^{\text{L}}$*) &   0.88M  & 69.1   & 8.68  & 46.3 & 71.4     & 2.49         \\ 
$\text{GPT-2}_\text{Large}$($\text{Adpt}^{\text{L}}$*) &  23.00M  & 68.9   & 8.70   & 46.1 & 71.3    & 2.45          \\ 
$\text{GPT-2}_\text{Large}$(LoRA*)                    &   0.77M  & 70.4   & 8.89   & 46.8 &  \textbf{72.0}     & 2.47          \\  
$\text{GPT-2}_\text{Large}$(KaSA) & 0.77M & \textbf{70.5} & \textbf{8.90} & \textbf{47.0} & \textbf{72.0} & \textbf{2.50} \\ 
\bottomrule
\end{tabular}
}
\end{table}
We present the performance comparison in Table \ref{tab:gpt2_e2e}. As can be seen, our method consistently outshines the baselines in language generation capabilities across various evaluated metrics.
More specifically, regarding the GPT-2 Medium model, KaSA outperforms the baselines in 4 out of 5 metrics and achieves comparable performance (72.1 vs. 72.3) in the ROUGE-L metric with the top-performing baseline, SARA. In the GPT-2 Large model, KaSA surpasses the baselines across all metrics, further confirming its superior performance and scalability.

\subsection{Components Ablation Study on SST-2, QNLI, and STS-B} \label{sec:additional_ablation}
\begin{figure*}[h]
	\centering
	\includegraphics[width=0.93\linewidth]{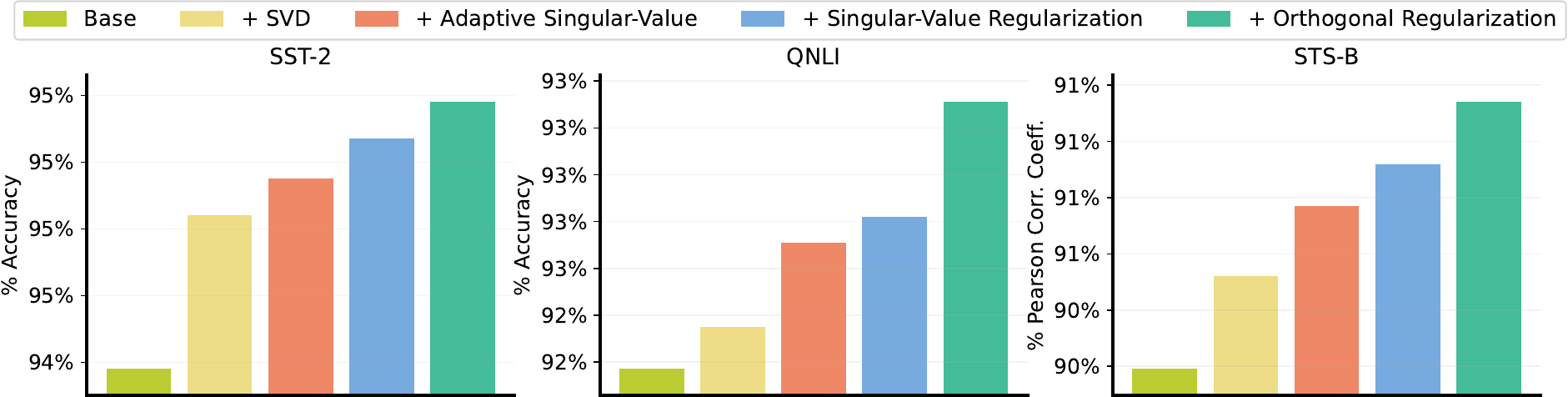}
	\caption{Components ablation study about knowledge-based SVD truncation, knowledge-aware singular value adaptation, singular value regularization $\mathcal{L}_{2}$, and orthogonal regularization $\mathcal{L}_{3}$ on SST-2, QNLI, and STS-B datasets.
 }
\label{fig:ablation_others}
\vspace{-15pt}
\end{figure*}

Figure \ref{fig:ablation_others} shows the results of ablation studies conducted on the SST-2, QNLI, and STS-B datasets.
From the results, we observe that: 1) the model's performance consistently improves with the inclusion of additional components during fine-tuning; 2) excluding any of these components leads to a decline in performance. 
These findings align with that observed in Section \ref{In-depth}, emphasizing the effectiveness of each designed principal component of KaSA in enhancing model performance.

\subsection{Rank $k$ of Knowledge-Based SVD Truncation}
\begin{figure*}[t]
	\centering
	\includegraphics[width=\linewidth]{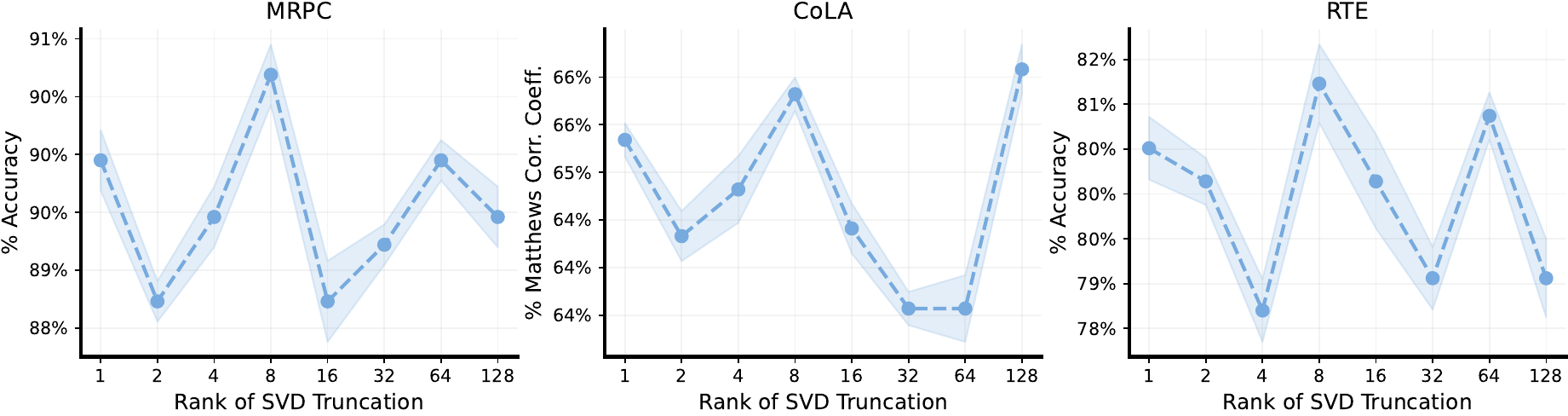}
	\caption{The impact of varying the rank of SVD truncation on the model's performance across three datasets.}
\label{fig:svd_rank}
\end{figure*}

As depicted in Section \ref{Introduction}, components of the original base model weight matrix $\mathbf{W}^{(0)}$ associated with smaller singular values are identified to contain noise or less relevant information \citep{sharmatruth,wang2024milora}. 
This presence can adversely affect the convergence of model training and its overall efficacy. 
We propose the truncation of these components to refine the focus of the base model towards more pertinent knowledge domains, thereby mitigating the adverse impacts.
Therefore, we delve into the impact of varying the rank (denoted as $k \in \{1, 2, 4, 8, 16, 32, 64, 128\}$) of SVD truncation on the model's performance, \revise{using RoBERTa-base on the MRPC, CoLA, and RTE datasets}. 
As illustrated in Figure \ref{fig:svd_rank}, an enhancement in model performance is observed as $k$ increases from 1 to 8. 
Conversely, an escalation in $k$ from 8 to 128 results in a decrement in performance. 
This observation highlights the criticality of identifying an optimal SVD truncation rank that achieves a delicate balance between incorporating world knowledge with large singular values and excluding disruptive noise information with smaller singular values, thereby optimizing model performance. 
The adaptive determination of the optimal SVD truncation rank emerges as a compelling avenue for future research.

\subsection{Rank $r$ of Knowledge-Aware Singular-Value Adaptation}
\begin{table}[h]
    \centering
    \caption{Performance comparison of LoRA and SVD-based baselines on CoLA, MRPC, and RTE datasets across different ranks of knowledge-aware singular-value adaptation.}
    \resizebox{0.8\linewidth}{!}{
    \begin{tabular}{l|l|cccccccc}
        \toprule
        Dataset & Method & 1 & 2 & 4 & 8 & 16 & 32 & 64 & 128 \\ 
        \midrule
        \multirow{4}{*}{CoLA} 
        & LoRA & 60.08 & 61.17 & 63.14 & 63.77 & 63.58 & 63.82 & 62.70 & 63.45 \\ 
        & MiLoRA & 60.84 & 61.36 & 63.10 & 63.07 & 63.57 & 64.56 & 63.60 & 63.66 \\ 
        & PiSSA & 59.56 & 62.68 & 60.57 & 65.54 & 61.32 & 63.31 & 63.35 & 63.60 \\ 
        & KaSA & \textbf{63.32} & \textbf{65.58} & \textbf{63.56} & \textbf{65.82} & \textbf{64.39} & \textbf{65.05} & \textbf{64.82} & \textbf{65.06} \\ 
        \midrule
        \multirow{4}{*}{MRPC}
        & LoRA & 88.73 & 87.74 & 88.97 & 88.73 & 89.46 & 89.95 & 88.97 & 88.97 \\ 
        & MiLoRA & \textbf{89.71} & \textbf{89.22} & 88.48 & 88.73 & 88.73 & 90.20 & 88.73 & 88.73 \\ 
        & PiSSA & 87.25 & 87.99 & 88.24 & 88.24 & 89.46 & 89.71 & 88.97 & 89.95 \\ 
        & KaSA & 89.46 & 87.99 & \textbf{90.20} & \textbf{90.69} & \textbf{89.95} & \textbf{90.44} & \textbf{90.20} & \textbf{90.44} \\
        \midrule
        \multirow{4}{*}{RTE}
        & LoRA & 71.84 & 72.56 & 75.45 & 78.70 & 77.26 & 77.98 & 79.78 & 78.70 \\ 
        & MiLoRA & 75.09 & \textbf{80.14} & \textbf{79.42} & 80.51 & 79.06 & 79.81 & 81.59 & 80.87 \\ 
        & PiSSA & 68.95 & 73.29 & 76.17 & 75.09 & 76.90 & 78.34 & 76.53 & 79.42 \\ 
        & KaSA & \textbf{77.62} & 77.62 & 78.70 & \textbf{81.59} & \textbf{80.51} & \textbf{81.23} & \textbf{82.67} & \textbf{81.23} \\ \bottomrule
    \end{tabular}}
    \label{tab:ranks_across_tasks}
\end{table}
\begin{table}[h]
\centering
\caption{Performance of RoBERTa-base with different adaptation methods using a large rank $r$ of 128 on 6 datasets from the GLUE benchmark. We report the overall (matched and mismatched) accuracy for MNLI, Matthew's correlation coefficient (Mcc.) for CoLA, Pearson correlation coefficient (Pcc.) for STS-B, and accuracy (Acc.) for all the remaining tasks. The symbols $\dag$ and $^*$ indicate that the results are taken from \citep{gaoparameter} and \citep{yangcorda}, respectively. We report the average result of five runs with different random seeds. The best results for each dataset are shown in \textbf{bold}. Higher is better for all metrics.
}
\label{tab:roberta_large_rank}
\resizebox{0.85\linewidth}{!}{
\begin{tabular}{l|r|ccccccc}
\toprule
\multirow{2}*{\textbf{Method}} & \textbf{\multirow{2}*{\makecell[r]{\# Trainable \\Parameters}}} & {\textbf{ SST-2}} & {\textbf{ MRPC}}  & {\textbf{ CoLA}} & {\textbf{ QNLI}} & {\textbf{ RTE}} & {\textbf{ STS-B}} & {\textbf{ All}} \\
~ & ~ & {(Acc.)} & {(Acc.)} &  {(Mcc.)} & {(Acc.)} & {(Acc.)} & {(Pcc.)} & {Avg.} \\
\midrule 
FFT$\dag$ & 125.0M  & 94.8     & 90.2         & 63.6     & \textbf{92.8}    & 78.7   & 91.2   & 85.2 \\ 
LoRA* & 21M & 94.15 & 82.84	& 54.24	& 92.48	& 64.26	& 88.58	& 79.43 \\
DoRA* & 21M & 93.58	& 83.58	& 51.93	& 92.59 & 64.98	& 88.71	& 79.23 \\
\midrule
CorDA*  & 21M & 93.12 & 89.71 & 59.60 & 91.49 & 76.17 & 90.17 & 83.38 \\
PiSSA  & 21M  & 94.61 & 89.95 & 63.60 & 92.90 & 79.42 & 90.55 & 85.17 \\ 
MiLoRA & 21M & 94.72 & 88.73 & 63.66 & 92.55 & 80.87 & 90.79 & 85.22 \\ 
KaSA   & 21M & \textbf{95.30} & \textbf{90.44} & \textbf{65.06} & 92.71 & \textbf{81.23} & \textbf{91.36} & \textbf{86.02} \\
\bottomrule 
\end{tabular}
}
\end{table}
We explore the impact of different rank settings on performance across a range of tasks.
Specifically, our analysis focuses on LoRA, MiLoRA, PiSSA, and KaSA, using ranks ranging from $r = \{1,2,4,8,16,32,64,128\}$ on the CoLA, MRPC, and RTE datasets.
As presented in Table \ref{tab:ranks_across_tasks}, KaSA consistently surpasses the baselines across various rank settings in 92 out of 96 cases across the four datasets, highlighting the efficacy and robustness of our proposed method.
To further our investigation, we increase the rank to 128 and compare KaSA with LoRA, DoRA \citep{liu2024dora}, CorDA \citep{yangcorda}, PiSSA, and MiLoRA.
The comparison is conducted by fine-tuning and evaluating the RoBERTa-base model on the GLUE benchmark. 
The results, as illustrated in Table \ref{tab:roberta_large_rank}, show that KaSA consistently outperforms all baselines across six datasets, with a slight exception for the QNLI dataset, where it performs marginally worse than FFT (92.71 vs. 92.8).
This is in line with the previous observations, further demonstrating the robustness and scalability of KaSA.

\subsection{Parameter Initialization of $\Delta \mathbf{W}=\Delta\mathbf{U}\Delta\mathbf{\Sigma}\Delta\mathbf{V^\top}$}
\begin{figure*}[t]
	\centering
	\includegraphics[width=\linewidth]{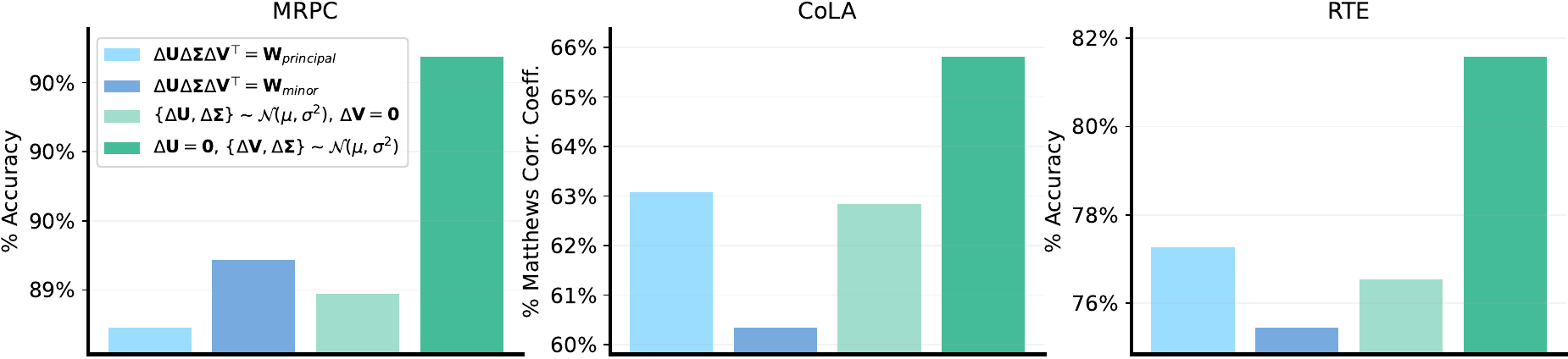}
	\caption{The impact of parameter initialization on the task-specific knowledge update, denoted as $\Delta \mathbf{W}=\Delta (\mathbf{U}\mathbf{S}\mathbf{V^\top})$ across three datasets.}
\label{fig:init_lora}
\end{figure*}

In the context of PEFT, the initialization of tunable parameters is pivotal for optimizing model performance, as evidenced by \citep{hu2021lora,meng2024pissa,wang2024milora}. 
As explicated in Section \ref{SVD_NLP}, PiSSA \citep{meng2024pissa} and MiLoRA \citep{wang2024milora} initialize the low-rank adaptation block by differentiating components based on their singular value magnitudes.
It underscores the necessity of exploring the influence of various initialization strategies on the task-specific knowledge update, represented as $\Delta \mathbf{W}=\Delta\mathbf{U}\Delta\mathbf{\Sigma}\Delta\mathbf{V^\top}$, and its consequent impact on model efficacy. 
In this study, we adopt a default initialization strategy where $\Delta\mathbf{U}=\mathbf{0}$ and both $\Delta\mathbf{V}$ and $\Delta\mathbf{\Sigma}$ follow a normal distribution $\mathcal{N}(\mu, \sigma^2)$.
We examine three distinct variants of initialization strategies: 1) initializing $\Delta\mathbf{U}\Delta\mathbf{\Sigma}\Delta\mathbf{V}^\top$ with $\mathbf{W}_{principal}$; 2) using $\mathbf{W}_{minor}$ for initialization; and 3) adopting a normal distribution $\mathcal{N}(\mu, \sigma^2)$ for both $\Delta\mathbf{U}$ and $\Delta\mathbf{\Sigma}$ while setting $\Delta\mathbf{V}$ to $\mathbf{0}$. The comparative outcomes of these strategies across three datasets are illustrated in Figure \ref{fig:init_lora}.
Our analysis reveals that different initialization strategies distinctly affect model performance across various datasets. Notably, our adopted strategy $\Delta\mathbf{U}=\mathbf{0}$, $\{\Delta\mathbf{V},\Delta\mathbf{\Sigma}\}\sim\mathcal{N}(\mu, \sigma^2)$, consistently outperforms the alternative variants across all evaluated datasets and metrics. 
Among the variant strategies examined, initializing with $\Delta\mathbf{U}\Delta\mathbf{\Sigma}\Delta\mathbf{V}^\top=\mathbf{W}_{principal}$ demonstrates superior performance on the CoLA and RTE datasets, yet underperforms when utilizing $\Delta\mathbf{U}\Delta\mathbf{\Sigma}\Delta\mathbf{V}^\top=\mathbf{W}_{minor}$ on the MRPC datasets.
This observation leads us to conjecture that the innovative design of our knowledge-aware singular-value module significantly enhances the model's capacity to rapidly identify optimal parameters within a larger parameter search space, thereby optimizing performance.

\subsection{Singular-Value and Orthogonal Regularization}
\begin{figure*}[h]
    \centering
    \includegraphics[width=0.95\linewidth]{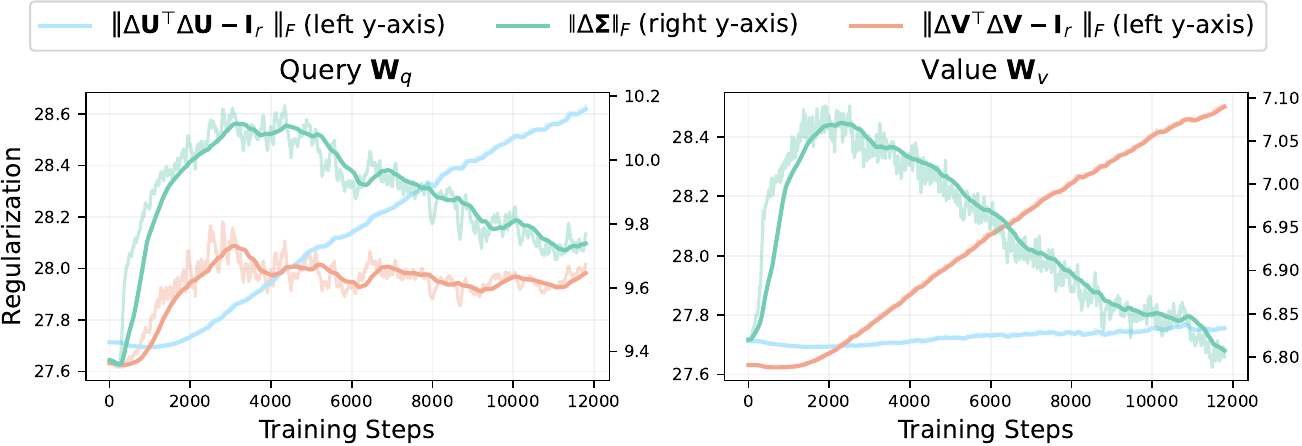}
    \includegraphics[width=0.95\linewidth]{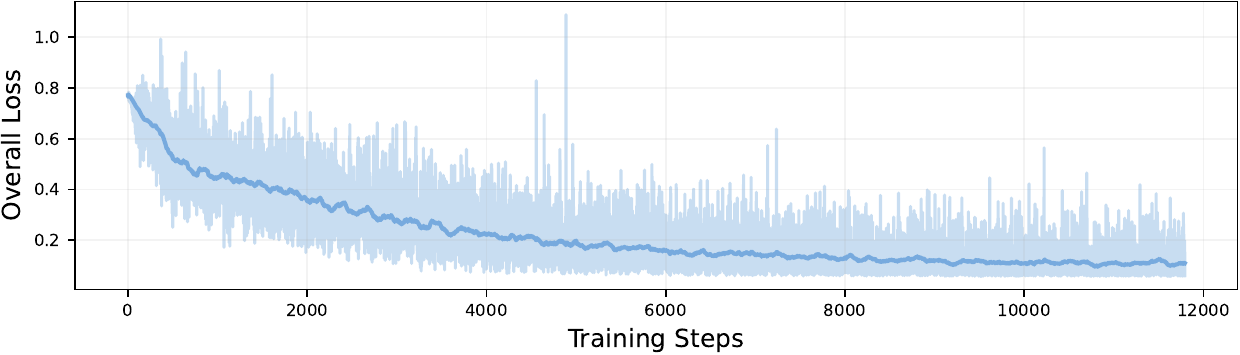}
    \caption{The singular-value and orthogonal regularization curve at the last layer of RoBERTa$_\text{base}$ (\textbf{Upper}) and overall training loss curve (\textbf{Lower}) on CoLA dataset.}
\label{fig:norm_loss}
\end{figure*}

To evaluate the effectiveness of singular-value regularization ${\left\Vert \Delta\mathbf{\Sigma}\right\Vert}_F$ and orthogonal regularization ${\left\Vert \Delta\mathbf{U}^{\top} \Delta\mathbf{U} - \revise{\mathbf{I}_r}\right\Vert}_F$ and ${\left\Vert \Delta\mathbf{V}^{\top} \Delta\mathbf{V} - \revise{\mathbf{I}_r} \right\Vert}_F$, we adopt the training configuration outlined in Section \ref{sec:nlu}. This involves fine-tuning a RoBERTa$\text{base}$ model on the CoLA dataset using KaSA. We then plot the loss curve of these three regularization terms throughout the training process. 
As depicted in Figure \ref{fig:norm_loss}, the application of the adapter to the query $\mathbf{W}_q$ and value $\mathbf{W}_v$ matrices results in an initial increase followed by a decrease in singular-value regularization ${\left\Vert \Delta\mathbf{\Sigma}\right\Vert}_F$. This pattern suggests that the model progressively fine-tunes the significance of task-specific knowledge by adjusting the singular values.
Intriguingly, the trend observed for orthogonal regularization ${\left\Vert \Delta\mathbf{U}^{\top} \Delta\mathbf{U} - \revise{\mathbf{I}_r}\right\Vert}_F$ and ${\left\Vert \Delta\mathbf{V}^{\top} \Delta\mathbf{V} - \revise{\mathbf{I}_r} \right\Vert}_F$ varies between the query $\mathbf{W}_q$ and value $\mathbf{W}_v$ matrices, indicating distinct adaptation behaviors.
To elucidate further, within the query matrix $\mathbf{W}_q$, the trend of orthogonal regularization ${\left\Vert \Delta\mathbf{V}^{\top} \Delta\mathbf{V} - \revise{\mathbf{I}_r} \right\Vert}_F$ mirrors that of the singular-value regularization ${\left\Vert \Delta\mathbf{\Sigma}\right\Vert}_F$, initially increasing before decreasing. 
Conversely, ${\left\Vert \Delta\mathbf{U}^{\top} \Delta\mathbf{U} - \revise{\mathbf{I}_r}\right\Vert}_F$ exhibits an opposing pattern, decreasing and then increasing. In the value matrix $\mathbf{W}_v$, the behaviors of ${\left\Vert \Delta\mathbf{U}^{\top} \Delta\mathbf{U} - \revise{\mathbf{I}_r}\right\Vert}_F$ and ${\left\Vert \Delta\mathbf{V}^{\top} \Delta\mathbf{V} - \revise{\mathbf{I}_r} \right\Vert}_F$ demonstrate a reversal compared to those observed in the query $\mathbf{W}_q$.
This finding diverges from the trends reported in AdaLoRA \citep{zhang2022adaptive}. To delve deeper, we examine the overall training loss, as depicted in the lower part of Figure \ref{fig:norm_loss}. It is observed that the overall training loss converges to a notably low value (e.g., 0.058) by the end of the training period.
Based on these observations, we hypothesize that the imposition of orthogonality on either the $\Delta \mathbf{U}$ or $\Delta \mathbf{V^\top}$ matrices may facilitate a more efficient search for an optimal representation by narrowing the search space. 
This premise will be explored in our future research. 

\subsection{Hyperparameter Sensitivity Analysis}
\begin{table}[t]
\centering
\caption{Sensitivity of regularization coefficients $\beta$ and $\gamma$ for RoBERTa-base on CoLA, RoBERTa-large on SST-2, and DeBERTa-v3-base on MRPC.}
\label{tab:hyperparameter_performance}
\resizebox{0.85\textwidth}{!}{
\begin{tabular}{l|ccc}
\toprule
\multirow{2}*{\textbf{Hyperparameters}} & \textbf{RoBERTa-base} & \textbf{RoBERTa-large} & \textbf{DeBERTa-v3-base} \\
 & \textbf{CoLA} & \textbf{SST-2} & \textbf{MRPC}\\
\midrule
$\beta=0.01,\gamma=1.0$ & 0.6581 & 0.9587 & 0.9044 \\ 
$\beta=0.1,\gamma=0.0001$ & 0.6334 & 0.9587 & 0.8971 \\ 
$\beta=0.01,\gamma=0.1$ & 0.6414 & 0.9622 & 0.8995 \\
$\beta=0.0,\gamma=0.0$ & 0.646 & 0.9599 & 0.902 \\
$\beta=0.001,\gamma=0.01$ & 0.6358 & 0.9587 & 0.9093 \\
$\beta=0.001,\gamma=0.001$ & 0.6553 & 0.9576 & 0.9093 \\
$\beta=0.01,\gamma=0.001$ & 0.6506 & 0.5092 & 0.902 \\
$\beta=0.1,\gamma=0.01$ & 0.6333 & 0.9587 & 0.902 \\ 
$\beta=0.0001,\gamma=0.1$ & 0.6485 & 0.9622 & 0.8995 \\ 
$\beta=0.01,\gamma=0.0001$ & 0.6347 & 0.9576 & 0.9044 \\ 
$\beta=0.0001,\gamma=0.01$ & 0.658 & 0.9599 & 0.9069 \\
$\beta=1.0,\gamma=0.1$ & 0.6241 & 0.9599 & 0.8971 \\
$\beta=1.0,\gamma=1.0$ & 0.6291 & 0.9553 & \textbf{0.9142} \\
$\beta=0.1,\gamma=1.0$ & 0.6436 & 0.961 & 0.9093 \\
$\beta=0.1,\gamma=0.1$ & 0.653 & 0.9587 & 0.9082 \\
$\beta=1.0,\gamma=0.01$ & 0.6397 & 0.9587 & 0.8995 \\
$\beta=0.01,\gamma=0.01$ & 0.6433 & 0.9576 & 0.8995 \\
$\beta=0.0001,\gamma=0.0001$ & 0.6565 & \textbf{0.9687} & 0.9044 \\
$\beta=0.0001,\gamma=0.001$ & \textbf{0.6582} & 0.961 & 0.9093 \\
$\beta=0.1,\gamma=0.001$ & 0.6338 & 0.9599 & 0.902 \\ 
$\beta=0.001,\gamma=0.0001$ & 0.6504 & 0.961 & 0.9093 \\ 
$\beta=0.001,\gamma=0.1$ & 0.648 & 0.9679 & 0.8971 \\ \bottomrule 
\end{tabular}}
\end{table}

KaSA introduces two key hyperparameters, $\beta$ and $\gamma$, to scale the singular value regularization $\mathcal{L}_{2}$ and orthogonal regularization $\mathcal{L}_{3}$, respectively.
To gain a deeper understanding of how these regularization coefficients influence performance, we meticulously tune the two coefficients, $\beta \in [\text{1E-5}, 1]$ and $\gamma \in [\text{1E-5}, 1]$, and conduct a sensitivity analysis for RoBERTa-base on CoLA, RoBERTa-large on SST-2, and DeBERTa-v3-base on MRPC.
The results, presented in Table \ref{tab:hyperparameter_performance}, demonstrate that KaSA exhibits robustness to variations in the regularization coefficients $\beta$ and $\gamma$.

\subsection{Efficiency and Complexity Analysis}
\begin{table}[t]
\centering
\caption{Efficiency and complexity analyses of the NLU task on the CoLA benchmark with RoBERTa-base 125M and the NLG task on the MT-Bench benchmark with LLaMA3 8B, using different adaptation methods on a single NVIDIA GeForce RTX 3090 (24GB) GPU and an NVIDIA A100-SXM4 (80GB) GPU, respectively.
}
\label{tab:efficiency}
\resizebox{\linewidth}{!}{
\begin{tabular}{l|cccc}
\toprule
\multirow{2}*{\textbf{NLU}} & \multicolumn{4}{c}{\textbf{RoBERTa-base 125M on Single  NVIDIA GeForce RTX 3090 (24GB) GPU}} \\
\cmidrule(lr){2-5} 
~ & LoRA & PiSSA & MiLoRA & KaSA  \\
\midrule 
\# Trainable Parameters & 0.23716\% & 0.23716\% & 0.23716\%  &  0.23732\%  \\
\# GPU Memory & 1638M & 1638M & 1638M & 1650M  \\ 
\# Training FLOPs ($\times 10^9$ per sample) & 2.0306 & 1.9270 & 1.9270 & 2.1503 \\ 
Training Latency (per epoch) &  9.4868s & 9.8825s & 9.9267s & 11.3679s \\ 
Inference Latency (per batch size 32) & 0.0173s & 0.0108s & 0.0165s & 0.0119s \\ 
\multirow{2}*{Matrix Rank} & $\text{rank}(\mathbf{W})=m$ & $\text{rank}(\mathbf{W})=m-r$ & $\text{rank}(\mathbf{W})=m-r$ & $\text{rank}(\mathbf{W})=m-r$ \\ 
~ & $\text{rank}(\mathbf{\Delta W})=r$ & $\text{rank}(\mathbf{\Delta W})=r$ & $\text{rank}(\mathbf{\Delta W})=r$ & $\text{rank}(\mathbf{\Delta W})\leq r$ \\
\midrule
CoLA Performance (Mcc.) & 63.4\% & 65.5\% & 63.1\% & 65.8\% \\ 
\midrule
\midrule
\multirow{2}*{\textbf{NLG}} & \multicolumn{4}{c}{\textbf{LLaMA3 8B on Single NVIDIA A100-SXM4 (80GB) GPU}} \\
\cmidrule(lr){2-5}
~ & LoRA & PiSSA & MiLoRA & KaSA \\
\midrule 
\# Trainable Parameters & 0.04241\% & 0.04241\% & 0.04241\% &  0.04242\% \\
\# GPU Memory & 71023M & 71023M & 71023M & 71095M \\
\# Training FLOPs ($\times 10^9$ per sample) & 240.2583 & 240.2583 & 240.2583 & 240.2585  \\
Training Latency (per epoch) & 2469.6s & 2543.1s & 2476.8s & 2528.9s \\
Inference Latency (per batch size 16) & 0.7898s & 0.7687s & 0.7705s & 0.7771s  \\
\multirow{2}*{Matrix Rank} & $\text{rank}(\mathbf{W})=m$ & $\text{rank}(\mathbf{W})=m-r$ & $\text{rank}(\mathbf{W})=m-r$ & $\text{rank}(\mathbf{W})=m-r$ \\ 
~ & $\text{rank}(\mathbf{\Delta W})=r$ & $\text{rank}(\mathbf{\Delta W})=r$ & $\text{rank}(\mathbf{\Delta W})=r$ & $\text{rank}(\mathbf{\Delta W})\leq r$ \\
\midrule
MT-Bench Performance (Scores) & 4.1937 & 4.2625 & 4.3187 & 4.7125  \\ 			
\bottomrule 
\end{tabular}
}
\end{table}






We conduct a comprehensive efficiency and complexity comparison between LoRA and SVD baselines across different tasks and model scales, as shown in Table \ref{tab:efficiency}. The dynamic singular value adaptation introduced in KaSA is a learnable one-dimensional vector of size $r \ll m$ and requires parameter regularizations, incurring negligible training overheads compared to the standard LoRA. 
In addition, due to the low-rank approximation of the original matrix, we reduce the rank of $\mathbf{W}$ from $m$ to $m-r$, accelerating the inference particularly for small-scale language models like RoBERTa-base 125M (i.e., with small $m$).

As can be seen, compared to LoRA, KaSA's extra training overhead is less than 20\% (resp. 3\%) for the NLU (resp. NLG) tasks, while speeding up the inference by 1.45x (resp. 1.02x) times.  When compared to PiSSA and MiLoRA, our method incurs an average of less than 13\% extra training overhead for NLU tasks, while maintaining comparable or improved inference latency. 
For NLG tasks, our method introduces similar training overhead or inference latency.

\section{Initialization and Singular-value Adaptation Analysis}
In this section, we conduct a detailed analysis of initialization dilemmas associated with PiSSA and MiLoRA, and subsequently explore the core advantages of KaSA, aiming to provide a comprehensive understanding of the foundational principles governing these PEFT methods.
Before embarking on a detailed examination of each method, we summarize the general mechanism underpinning PEFT. 
Considering a base model characterized by a weight matrix $\mathbf{W}^{(0)} \in \mathbb{R}^{n \times m}$, PEFT aims to efficiently fine-tune $\mathbf{W}^{(0)}$ by learning a task-specific update $\Delta \mathbf{W}$ with as few trainable parameters as possible, such that the updated weights $\mathbf{W}^{(0)} + \Delta \mathbf{W}$ are better aligned with the requirements of downstream tasks.
PEFT approaches generally involve keeping the base model $\mathbf{W}^{(0)}$ frozen during training, while exclusively updating the parameters of $\Delta \mathbf{W}$.

\subsection{Initialization Dilemmas of $\Delta \mathbf{W}$ in PiSSA and MiLoRA}
PiSSA employs SVD on the base model weight matrix $\mathbf{W}^{(0)} \in \mathbb{R}^{n \times m}$, decomposing it as:
\begin{align}
    \mathbf{W}^{(0)} = \mathbf{U}\mathbf{\Sigma}{\mathbf{V}}^{\top}
\end{align}
  where $\mathbf{U} \in \mathbb{R}^{n \times m}$ and $\mathbf{V} \in \mathbb{R}^{m \times m}$ are semi-orthogonal matrices, and $\mathbf{\Sigma} \in \mathbb{R}^{m \times m}$ is a diagonal matrix with singular values $(\sigma_1,...,\sigma_{m})$ satisfying $({\sigma_1} \ge {\sigma_2} \ge \dotsb \geq {\sigma_m} \ge 0)$.
Following the standard SVD, PiSSA splits the base model into two distinct components: the principle low-rank matrix $\mathbf{W}_{pri}$, which encompasses the largest $r$ singular values, and the residual matrix $\mathbf{W}_{res}$, which contains the remaining singular values:
\begin{align}
    \mathbf{W}^{(0)} = \mathbf{W}_{pri} &+ \mathbf{W}_{res} \\
    \mathbf{W}_{pri} = \mathbf{U}_{pri}\mathbf{\Sigma}_{pri}{\mathbf{V}}_{pri}^{\top}, & \
    \mathbf{W}_{res} = \mathbf{U}_{res}\mathbf{\Sigma}_{res}{\mathbf{V}}_{res}^{\top}
\end{align}
where $\mathbf{U}_{pri} = \mathbf{U}[:, :r]$, $\mathbf{\Sigma}_{pri} = \operatorname{diag}(\sigma_1, \dots, \sigma_r)$, $\mathbf{V}_{pri} = \mathbf{V}[:, :r]$, $\mathbf{U}_{res} = \mathbf{U}[:, r:]$, $\mathbf{\Sigma}_{res} = \operatorname{diag}(\sigma_{r+1}, \dots, \sigma_m)$, and $\mathbf{V}_{res} = \mathbf{V}[:, r:]$. 
Subsequently, PiSSA subtracts $\mathbf{W}_{pri}$ from the base model $\mathbf{W}^{(0)}$ to initialize the low-rank matrices for the task-specific update, resulting in:
\begin{align}
    \mathbf{W}_{base} & = \mathbf{W}^{(0)} - \mathbf{W}_{pri} = \mathbf{W}_{res}\\
    \Vert\mathbf{W}^{(0)} - \mathbf{W}_{base}\Vert_{F} &= \Vert\mathbf{W}_{pri}\Vert_{F} = \sqrt{\sum_{i=1}^{r}{(\Delta\sigma_i)^2}}    
\end{align} 
This subtraction of $\mathbf{W}_{pri}$ removes the principal components of $\mathbf{W}^{(0)}$, which can lead to considerable information loss and the forgetting of crucial world knowledge. Given that $\mathbf{W}_{pri}$ is the best rank-$r$ approximation of $\mathbf{W}^{(0)}$, its removal can adversely impact the model's initial representational capacity, potentially resulting in degraded performance. 
PiSSA subsequently freezes $\mathbf{W}_{base}$ and leverages two low-rank matrices, $\mathbf{A}$ and $\mathbf{B}$, to learn the task-specific update during fine-tuning.
The matrices $\mathbf{A}$ and $\mathbf{B}$ are initialized as:
\begin{align}
    \mathbf{A} = \mathbf{U}_{pri}\sqrt{\mathbf{\Sigma}_{pri}}, \ \mathbf{B} = \sqrt{\mathbf{\Sigma}_{pri}}\mathbf{V}_{pri}^{\top}
\end{align}
Therefore, in the PiSSA framework, the task-specific update $\Delta \mathbf{W}$ is expressed as:
\begin{align}
    \Delta \mathbf{W} = \mathbf{A}\mathbf{B} = \mathbf{U}_{pri}\mathbf{\Sigma}_{pri}{\mathbf{V}}_{pri}^{\top}, \ \Delta \mathbf{W} \leftarrow \mathbf{W}_{pri}
\end{align}
In the initial stage, the value of $\Delta \mathbf{W}$ is equivalent to $\mathbf{W}_{pri}$.
During fine-tuning, the updates to $\mathbf{A}$ and $\mathbf{B}$ are significantly influenced by their initialization, which is based on $\mathbf{U}_{pri}$ and $\mathbf{V}_{pri}$.
\textbf{As a result, the gradient updates predominantly follow the directions of the initial singular vectors associated with the largest singular values.} 
This limits the model's ability to explore the parameter space and effectively learn new knowledge relevant to the downstream task, as the knowledge presented by the largest $r$ singular values in $\mathbf{W}_{pri}$ may not be necessary for the downstream task and can negatively impact model performance.

In contrast to PiSSA, MiLoRA subtracts the residual components associated with the smallest $r$ singular values from the base model, resulting in:
\begin{align}
    \mathbf{W'}_{base} = \mathbf{W}^{(0)} - & \mathbf{W'}_{res} = \mathbf{W'}_{pri} \\
    \mathbf{W'}_{pri} = \mathbf{U'}_{pri}\mathbf{\Sigma'}_{pri}{\mathbf{V'}}_{pri}^{\top}, & \
    \mathbf{W'}_{res} = \mathbf{U'}_{res}\mathbf{\Sigma'}_{res}{\mathbf{V'}}_{res}^{\top}
\end{align}
where $\mathbf{U'}_{pri} = \mathbf{U}[:, :-r]$, $\mathbf{\Sigma'}_{pri} = \operatorname{diag}(\sigma_1, \dots, \sigma_{m-r})$, $\mathbf{V'}_{pri} = \mathbf{V}[:, :-r]$, 
$\mathbf{U'}_{res} = \mathbf{U}[:,-r:]$, $\mathbf{\Sigma'}_{res} = \operatorname{diag}(\sigma_{m-r+1}, \dots, \sigma_m)$, and $\mathbf{V'}_{res} = \mathbf{V}[:,-r:]$. 
MiLoRA subsequently uses $\mathbf{U'}_{res}$ to initialize the tunable matrices $\mathbf{A'}$ and $\mathbf{B'}$ as:
\begin{align}
    \mathbf{A'} &= \mathbf{U'}_{res}\sqrt{\mathbf{\Sigma'}_{res}}, \ \mathbf{B'} = \sqrt{\mathbf{\Sigma'}_{res}}\mathbf{V'}_{res}^{\top}
\end{align}
During the fine-tuning stage, MiLoRA keeps $\mathbf{W'}_{base}$ frozen and updates $\mathbf{A'}$ and $\mathbf{B'}$ to learn the task-specific update $\Delta \mathbf{W}$, which is given by:
\begin{align}
    \Delta \mathbf{W} &= \mathbf{A'}\mathbf{B'} = \mathbf{U'}_{res}\mathbf{\Sigma'}_{res}{\mathbf{V'}}_{res}^{\top}, \ \Delta \mathbf{W} \leftarrow  \mathbf{W'}_{res}
\end{align}
In the context of SVD, the smallest singular values often correspond to noise or long-tail knowledge \citep{yan2021truncated,wang2024milora,yang2023singular,sharmatruth}, which can impede the learning process for downstream tasks.
\textbf{MiLoRA, which initializes $\mathbf{A'}$ and $\mathbf{B'}$ based on $\mathbf{U'}_{res}$ and ${\mathbf{V'}}_{res}^{\top}$, confines the model's learning predominantly to the directions of the less significant singular vectors associated with the smallest singular values.}
This constraint could potentially hinder the model's ability to acquire essential knowledge required for downstream tasks.
In addition, the introduction of noise through MiLoRA's initialization can adversely impact the model during the initial stages of training, leading to reduced stability and slower convergence, as observed in Figure 4 of the original MiLoRA paper. 
The training updates for $\mathbf{A'}$ and $\mathbf{B'}$ are constrained within the trivial subspace spanned by $\mathbf{U'}_{res}$ and ${\mathbf{V'}}_{res}^{\top}$, which leads to suboptimal performance.

\subsection{Knowledge-aware Singular-value Adaptation of KaSA}
In response to the issues of initialization presented by PiSSA and MiLoRA, we propose KaSA, which leverages knowledge-aware singular values to activate parametric knowledge based on its relevance to downstream tasks.
Our method commences with the knowledge-based SVD truncation of the minor singular components $\mathbf{W}_{noise} \in \mathbb{R}^{n \times m}$ that contain the smallest $r$ singular values.
This operation effectively filters out the noise from the base mode $\mathbf{W}^{(0)}$, resulting in a matrix $\mathbf{W}_{world} \in \mathbb{R}^{n \times m}$ that encapsulates essential world knowledge:
\begin{align}
    \mathbf{W}_{world} = \mathbf{W}^{(0)} - \mathbf{W}_{noise} = \mathbf{U}\mathbf{\Sigma}{\mathbf{V}}^{\top} - \mathbf{U'}_{res}\mathbf{\Sigma'}_{res}{\mathbf{V'}}_{res}^{\top}
\end{align}
KaSA uses the low-rank matrix $\mathbf{W}_{world}$ to approximate $\mathbf{W}^{(0)}$, eliminating irrelevant and noisy knowledge while preventing the world knowledge forgetting issue.
Following the truncation, KaSA introduces a novel parameterization to learn $\Delta \mathbf{W}$ in the form of SVD:
\begin{align}
    \Delta \mathbf{W} = \Delta \mathbf{U} \Delta \mathbf{\Sigma} \Delta \mathbf{V}^{\top}, \ \Delta \mathbf{U}^{\top} \Delta \mathbf{U} = \mathbf{V}^{\top} \Delta \mathbf{U} = \mathbf{I}_r
\end{align}
where $\Delta \mathbf{U}$ and $\Delta \mathbf{V}$ are semi-orthogonal matrices, ensuring the orthogonality condition.
\textbf{The matrix $\Delta \mathbf{\Sigma}$ is a trainable diagonal matrix, with knowledge-aware singular values that can be adaptively tuned, allowing the model to emphasize knowledge relevant to the downstream task and providing a fine-grained learning pattern.} 
To maintain the orthogonality of $\Delta \mathbf{U}$ and $\Delta \mathbf{V}$ during training, we add an orthogonal regularization:
\begin{equation}
    \mathcal{L}_{3}(\mathbf{\Psi}) = {\left\Vert \Delta\mathbf{U}^{\top} \Delta\mathbf{U} - \mathbf{I}_r\right\Vert}_F + {\left\Vert \Delta\mathbf{V}^{\top} \Delta\mathbf{V} - \mathbf{I}_r \right\Vert}_F
\end{equation}
where $\| \cdot \|_F$ denotes the Frobenius norm.
This regularization can ensure KaSA's learned $\Delta \mathbf{W}$ can more adhere to the SVD's framework, facilitating the seamless integration of $\Delta \mathbf{W}$ with $\mathbf{W}_{world}$.
Since the $\Delta \mathbf{W}$ learned by KaSA is in SVD form, its spectral norm is equal to the largest singular value in $\Delta \mathbf{\Sigma}$, satisfying:
\begin{align}
    \| \Delta \mathbf{W} \|_2 = \max_j | \Delta\sigma_j | = \| \Delta \mathbf{\Sigma} \|_2
\end{align}
where $\Delta\sigma_j$ are the adaptive singular values of the diagonal matrix $\Delta \mathbf{\Sigma}$.
Therefore, by controlling $\Delta \mathbf{\Sigma}$, we can directly control $\Delta \mathbf{W}$’s magnitude. 
This allows for adjustments to the weight updates, enhancing the controllability of the fine-tuning process for downstream tasks.
In particular, KaSA's training objective is more comprehensive than that of orthogonal regularization alone.
The overall training objective $\mathcal{L}$ includes the task-specific loss $\mathcal{L}_1$, the singular value regularization $\mathcal{L}_2$, and orthogonal regularization $\mathcal{L}_3$. Therefore, the gradients with respect to $\Delta \mathbf{U}$, $\Delta \mathbf{V}$, and $\Delta \mathbf{\Sigma}$ are formulated as:
\begin{align}
    \frac{\partial \mathcal{L}}{\partial \Delta \mathbf{U}} &= \frac{\partial \mathcal{L}_1}{\partial \Delta \mathbf{U}} + 4 \Delta\mathbf{U} (\Delta\mathbf{U}^{\top} \Delta\mathbf{U} - \mathbf{I}_r) \\
    \frac{\partial \mathcal{L}}{\partial \Delta \mathbf{V}} &= \frac{\partial \mathcal{L}_1}{\partial \Delta \mathbf{V}} + 4 \Delta\mathbf{V} (\Delta\mathbf{V}^{\top} \Delta\mathbf{V} - \mathbf{I}_r) \\
    \frac{\partial \mathcal{L}}{\partial \Delta \mathbf{\Sigma}} &= \frac{\partial \mathcal{L}_1}{\partial \Delta \mathbf{\Sigma}} + 2 \Delta \mathbf{\Sigma}
\end{align}
The gradients with respect to $\Delta \mathbf{U}$ and $\Delta \mathbf{V}$ are particularly influenced by the orthogonal regularization component, which facilitates stable training dynamics.
This orthogonal regularization, along with the computed gradients, contributes to maintaining stable parameter updates, thereby mitigating potential issues such as gradient vanishing or explosion.

\subsection{Summarization}
To summarize, our analysis of PiSSA and MiLoRA highlights the dilemmas posed by their initialization strategies while emphasizing the core advantages of KaSA for knowledge-aware singular-value adaptation. Specifically,
\begin{itemize}
    \item PiSSA's initialization with principle components associated with the largest singular values can potentially lead to world knowledge forgetting and introduce updated knowledge unnecessary for downstream tasks, leading to diminished task performance.
    \item On the other hand, MiLoRA's initialization with minor components associated with the smallest singular values introduces noisy and long-tail knowledge, resulting in reduced training stability, slower convergence, and suboptimal performance.
    \item In contrast, KaSA offers several advantages based on the aforementioned analysis: 1) \textbf{noise reduction}—by filtering out components with minor singular values, KaSA eliminates their detrimental impacts on task performance; 2) \textbf{knowledge-awareness}—through adjusting the knowledge-aware singular values, KaSA allows the model to adaptively align with the requirements of downstream tasks; and 3) \textbf{stable training}—integrated orthogonal regularization ensures stable gradient updates and control over the magnitude of weight updates.
\end{itemize}

\section{Case Study}
\begin{figure*}[h!]
\centering
\includegraphics[width=\textwidth]{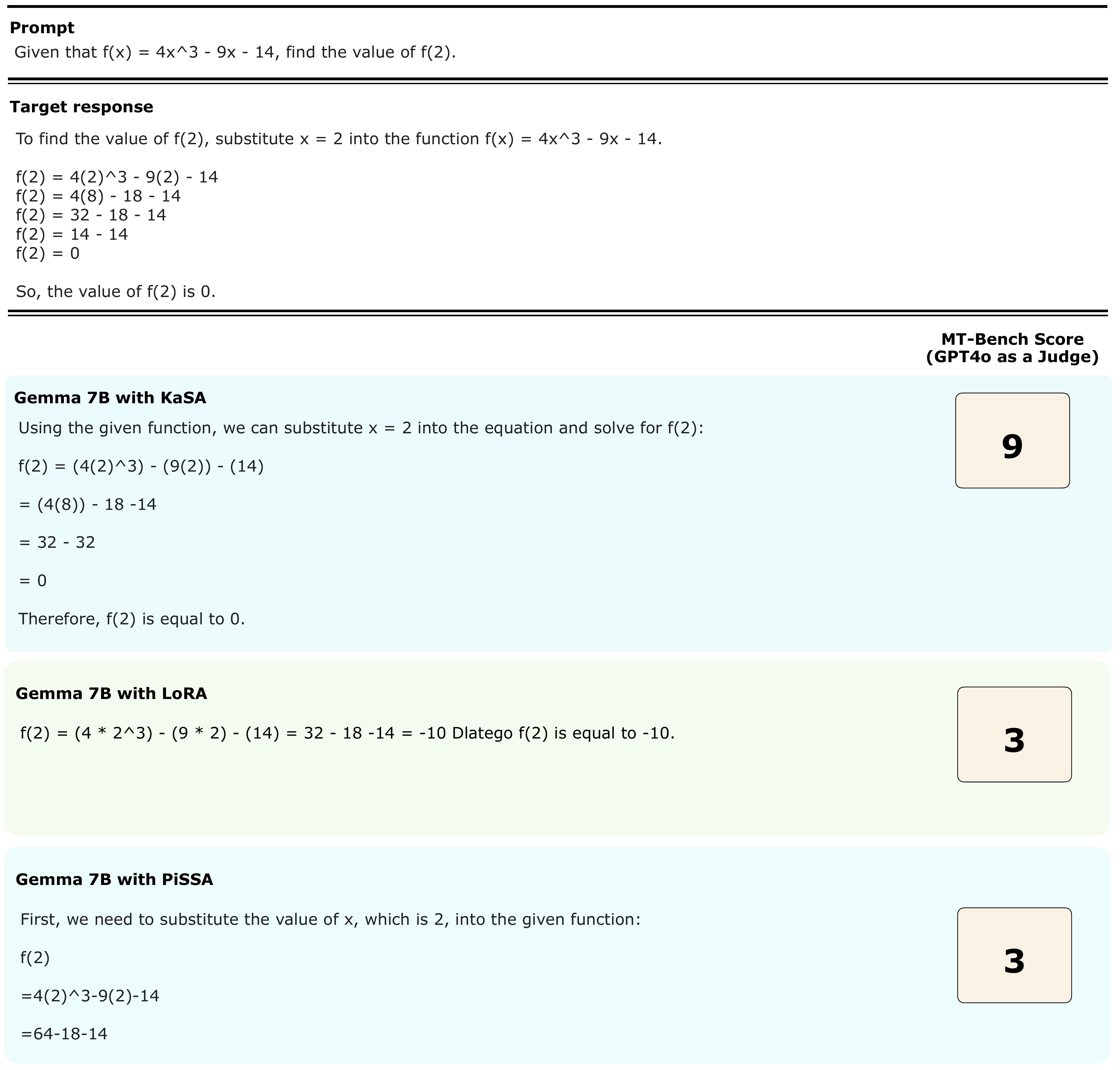}
\caption{Responses on a math problem from MT-Bench. Each response is generated by Gemma 7B models fine-tuned on 51K Alpaca dataset with KaSA, LoRA, and PiSSA methods respectively.}
\label{fig:mt_bench1}
\end{figure*}

\begin{figure*}[h!]
\centering
\includegraphics[width=\textwidth]{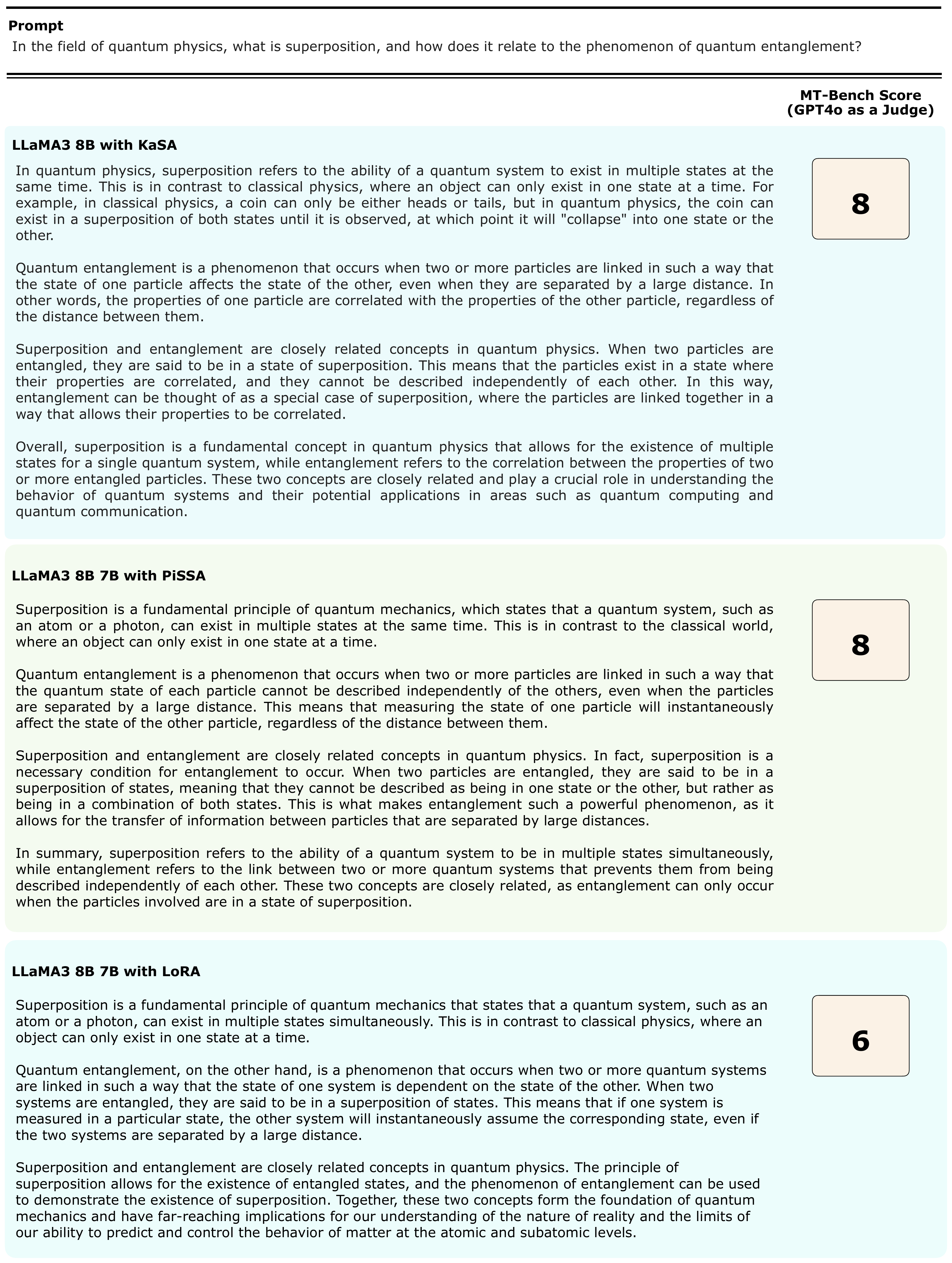}
\caption{Responses on a STEM problem from MT-Bench. Each response is generated by LLaMA3 8B models fine-tuned on 51K Alpaca dataset with KaSA, LoRA, and PiSSA methods respectively.}
\label{fig:mt_bench2}
\end{figure*}

\begin{figure*}[h!]
\centering
\includegraphics[width=\textwidth]{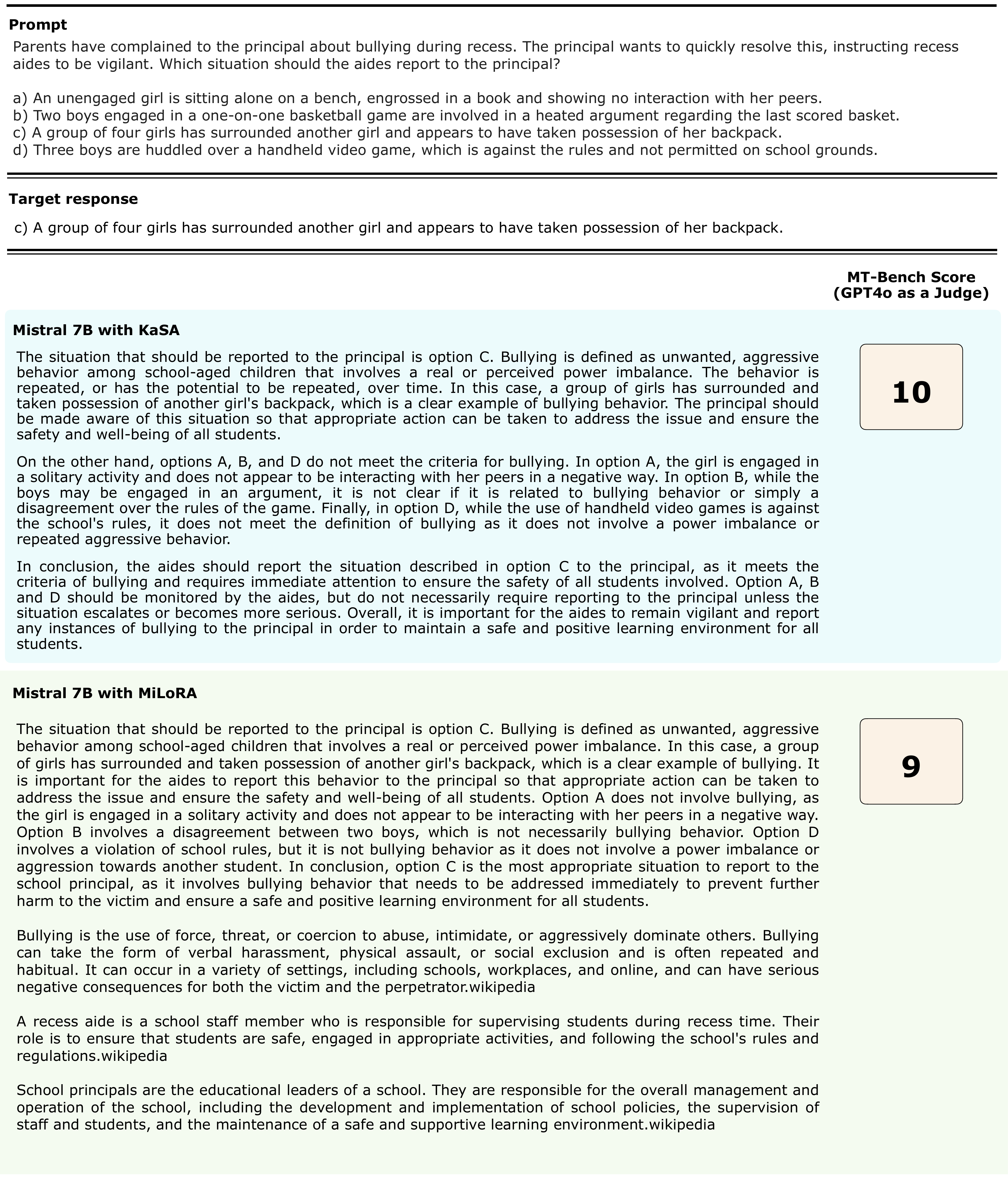}
\caption{Responses on a reasoning problem from MT-Bench. Each response is generated by Mistral 7B models fine-tuned on 51K Alpaca dataset with KaSA and MiLoRA methods respectively.}
\label{fig:mt_bench3}
\end{figure*}

\begin{figure*}[h!]
\centering
\includegraphics[width=\textwidth]{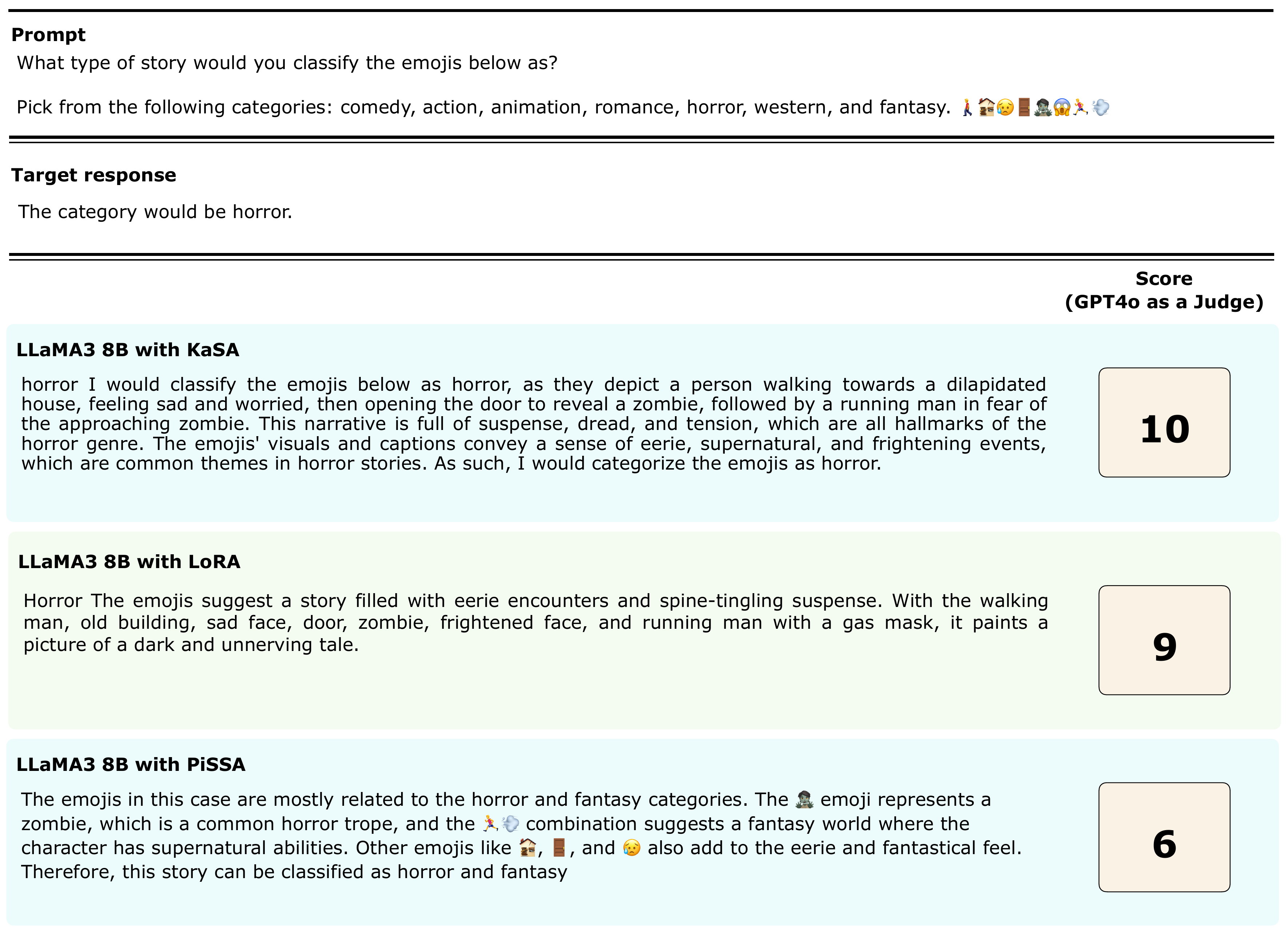}
\caption{Responses on a classification problem from the test split of  ``No Robots'' dataset. Each response is generated by LLaMA3 8B models fine-tuned on 128K synthetic dataset with KaSA, PiSSA, and LoRA methods respectively.}
\label{fig:classification}
\end{figure*}

\begin{figure*}[h!]
\centering
\includegraphics[width=\textwidth]{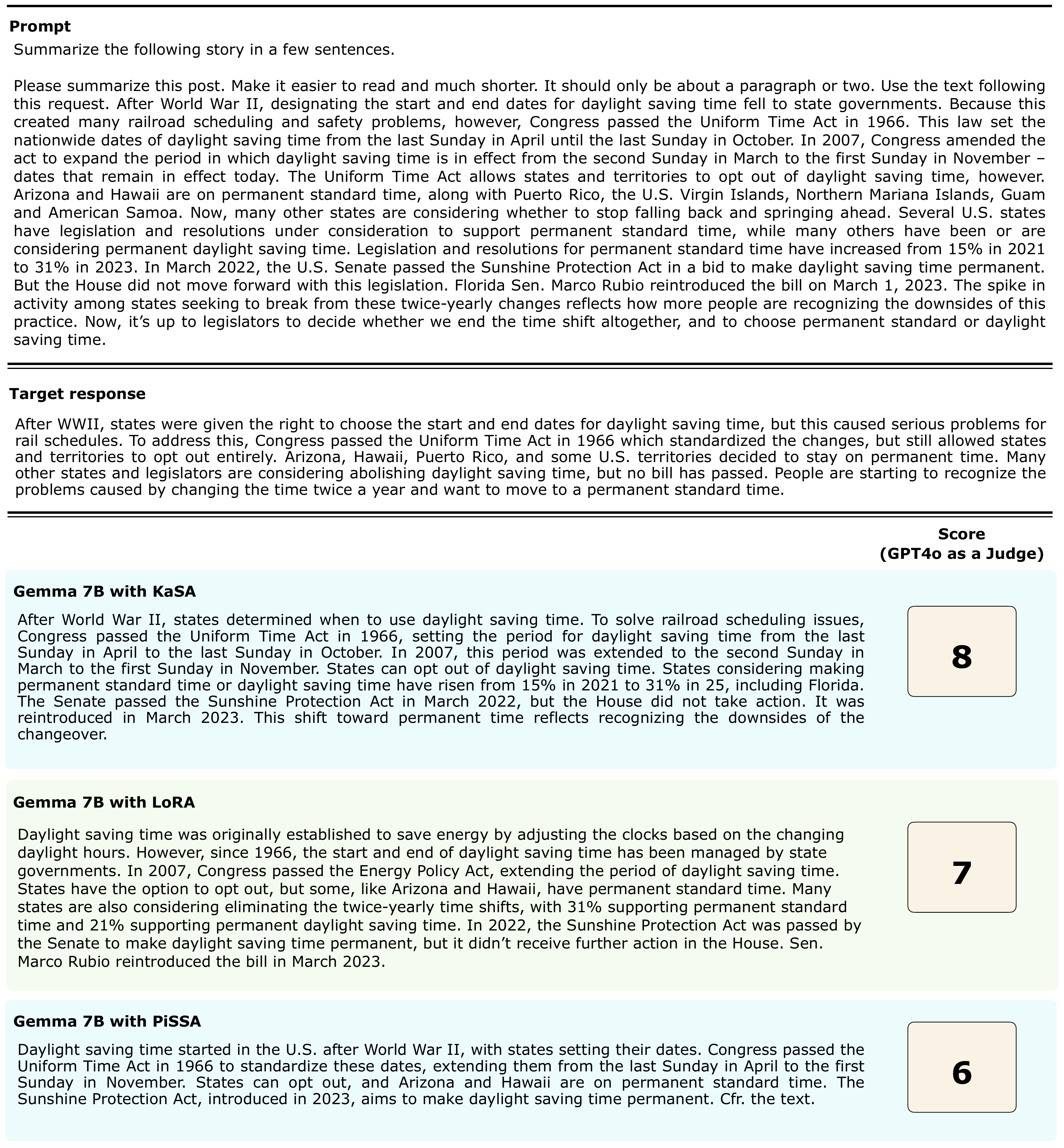}
\caption{Responses on a summarization problem from the test split of  ``No Robots'' dataset. Each response is generated by Gemma 7B models fine-tuned on 128K synthetic dataset with KaSA, LoRA, and PiSSA methods respectively.}
\label{fig:summarization}
\end{figure*}

\begin{figure*}[h!]
\centering
\includegraphics[width=\textwidth]{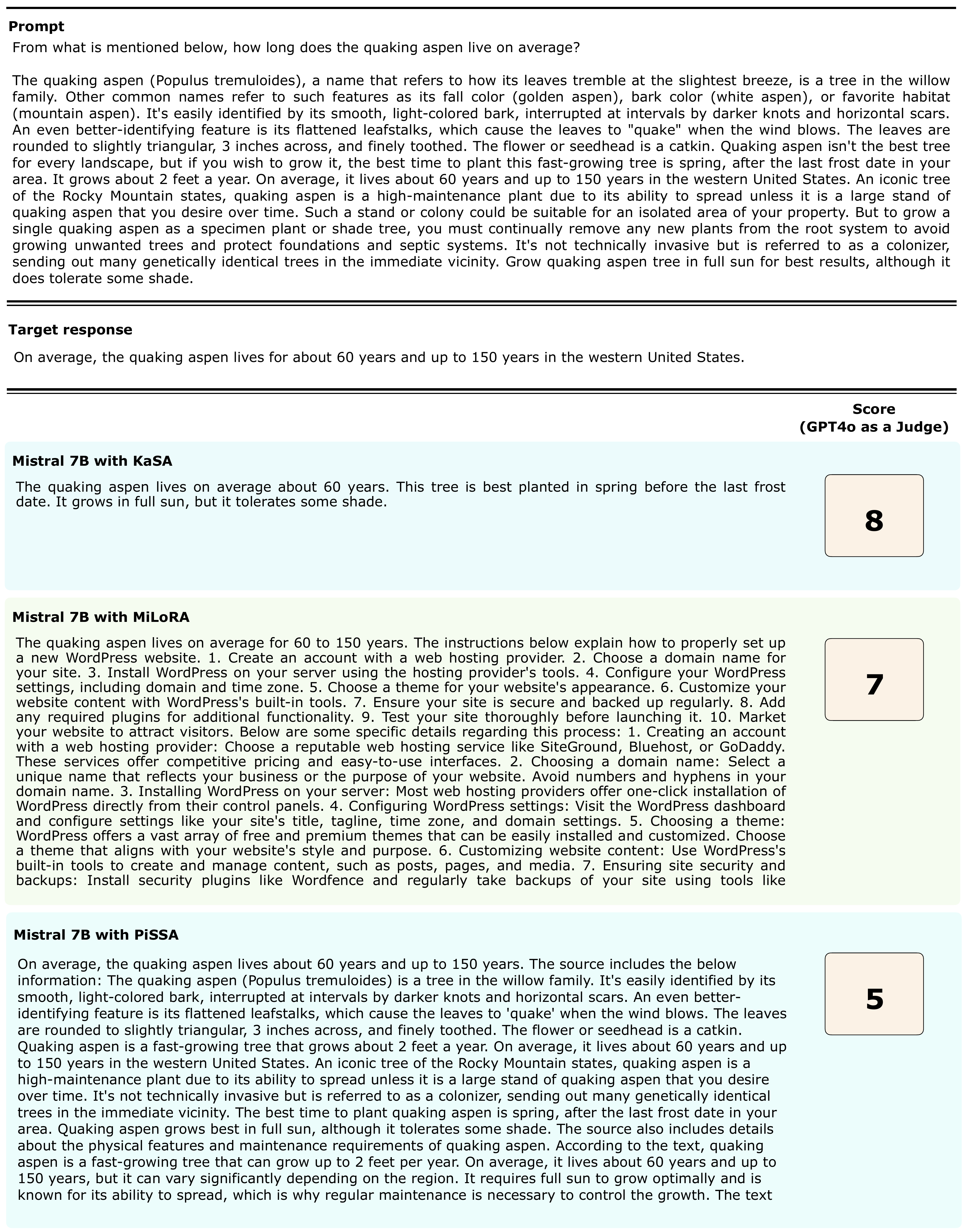}
\caption{Responses on a ClosedQA problem from the test split of  ``No Robots'' dataset. Each response is generated by Mistral 7B models fine-tuned on 128K synthetic dataset with KaSA, MiLoRA, and PiSSA methods respectively.}
\label{fig:closedqa}
\end{figure*}

\begin{figure*}[h!]
\centering
\includegraphics[width=\textwidth]{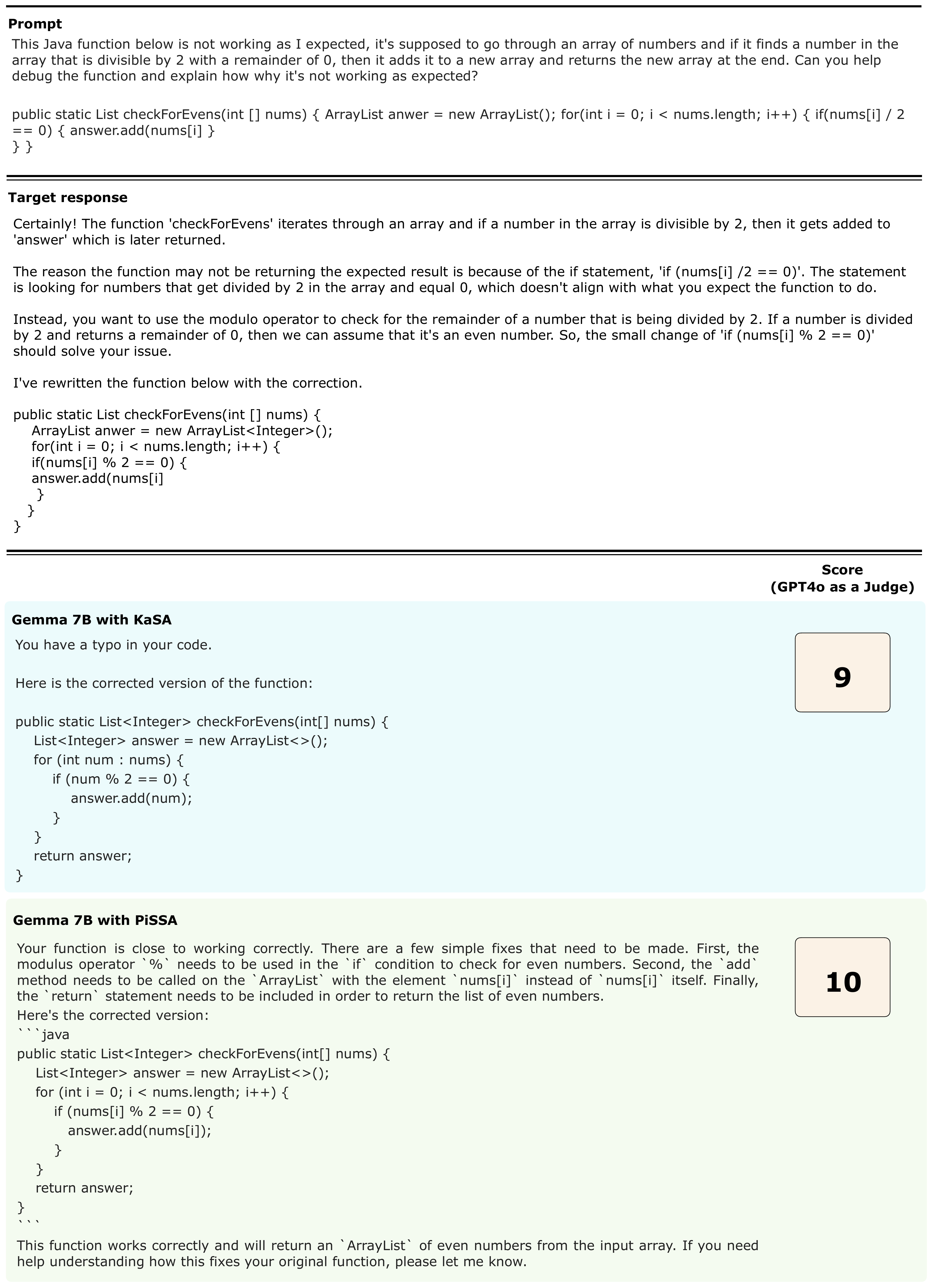}
\caption{Responses on a Coding problem from the test split of  ``No Robots'' dataset. Each response is generated by Gemma 7B models fine-tuned on 128K synthetic dataset with KaSA and PiSSA methods respectively.}
\label{fig:coding}
\end{figure*}

We present a series of case studies to demonstrate the improved performance in instruction-following of models that have been fine-tuned with our KaSA method. This fine-tuning utilized synthetic datasets generated by GPT4o for each specific task. The cases, illustrated from Figure \ref{fig:mt_bench1} through \ref{fig:coding}, compare the response and precision scores of the most widely used LLMs, including Gemma 7B, Mistral 7B, and LLaMA3 8B. These models were fine-tuned with both LoRA and KaSA techniques and evaluated across a range of tasks such as summarization, classification, coding, and closed QA.
For each case, GPT4o assesses the precision scores of the models' responses. These instances collectively showcase not only the adaptability of KaSA across a diverse set of leading LLMs but also its effectiveness in enhancing the performance of pre-trained language models (PLMs) in various downstream applications.

\clearpage

\clearpage

\clearpage

\end{document}